%% file: omniwise.tex
\documentclass[sigconf,nonacm]{acmart}

\input{tex/package}
\input{tex/macros}

\AtBeginDocument{%
  \providecommand\BibTeX{{%
    \normalfont B\kern-0.5em{\scshape i\kern-0.25em b}\kern-0.8em\TeX}}}

\begin{document}

\title{\omniwise: Predicting GPU Kernels Performance with LLMs}

\author{Zixian Wang}
\email{zixianw4@illinois.edu}
\orcid{0009-0007-7123-2908}
\affiliation{%
  \institution{University of Illinois Urbana-Champaign}
  \city{Urbana, IL}
  \country{USA}}
\authornote{Work done as an intern at AMD.}

\author{Cole Ramos}
\email{Cole.Ramos@amd.com}
\affiliation{%
  \institution{AMD}
  \city{Austin, TX}
  \country{USA}
}

\author{Muhammad A. Awad}
\email{muhaawad@amd.com}
\orcid{0000-0002-6914-493X}
\affiliation{%
  \institution{AMD}
  \city{Santa Clara, CA}
  \country{USA}
}

\author{Keith Lowery}
\email{Keith.Lowery@amd.com}
\affiliation{%
  \institution{AMD}
  \city{Austin, TX}
  \country{USA}
}

\renewcommand{\shortauthors}{Wang, Ramos, Awad, and Lowery}

\input{tex/abstract}


\maketitle
\input{tex/intro}

\input{tex/background}

\input{tex/omniwise}
\input{tex/results}

\input{tex/conclusion}

\input{tex/acknowledgments}

\bibliographystyle{ACM-Reference-Format}
\bibliography{temp}

\end{document}

%% file: tex/package.tex
\usepackage{algorithmic}
\usepackage{hyperref}
\usepackage{graphicx}
\usepackage{textcomp}
\usepackage{xcolor}
\usepackage{subcaption}
\usepackage{booktabs}
\usepackage{natbib}
\usepackage{multirow}
\usepackage{balance}
\usepackage{soul, listings,lstautogobble}
\usepackage{ color, colortbl}
\usepackage[normalem]{ulem} 
\usepackage{float}
\usepackage{xspace}
\usepackage{geometry}
\usepackage{paralist}

\definecolor{ListBGColor}{rgb}{0.95,0.95,0.95}
\definecolor{KeywordColor}{rgb}{0,0,0.6}
\definecolor{ListCommentColor}{rgb}{0.4, 0.27, 0.173}
\makeatletter
\newcommand\labelline[1]{%
  \def\@currentlabel{\thelstnumber}\label{#1}}
\newcommand\lineref[1]{
  \@ifundefined{r@#1}{0}{\ref{#1}}
}
\makeatother

%% file: tex/macros.tex
\makeatletter
\providecommand*{\NAT@spacechar}{~}
\makeatother

\newcommand{\lsref}[1]{Listing~\ref{lst:#1}}
\newcommand{\secref}[1]{Section~\ref{sec:#1}}
\newcommand{\figref}[1]{Figure~\ref{fig:#1}}
\newcommand{\tabref}[1]{Table~\ref{tab:#1}}

\newcommand{\omniwise}{Omniwise\xspace}
\newcommand{\llama}{\textsc{LLaMA}\xspace}

\colorlet{punct}{red!60!black}
\definecolor{background}{HTML}{EEEEEE}
\definecolor{delim}{RGB}{20,105,176}
\colorlet{numb}{magenta!60!black}

\lstdefinelanguage{json}{
    basicstyle=\normalfont\ttfamily,
    numbers=left,
    numberstyle=\scriptsize,
    stepnumber=1,
    numbersep=8pt,
    showstringspaces=false,
    breaklines=true,
    frame=lines,
    backgroundcolor=\color{background},
    literate=
     *{0}{{{\color{numb}0}}}{1}
      {1}{{{\color{numb}1}}}{1}
      {2}{{{\color{numb}2}}}{1}
      {3}{{{\color{numb}3}}}{1}
      {4}{{{\color{numb}4}}}{1}
      {5}{{{\color{numb}5}}}{1}
      {6}{{{\color{numb}6}}}{1}
      {7}{{{\color{numb}7}}}{1}
      {8}{{{\color{numb}8}}}{1}
      {9}{{{\color{numb}9}}}{1}
      {:}{{{\color{punct}{:}}}}{1}
      {,}{{{\color{punct}{,}}}}{1}
      {\{}{{{\color{delim}{\{}}}}{1}
      {\}}{{{\color{delim}{\}}}}}{1}
      {[}{{{\color{delim}{[}}}}{1}
      {]}{{{\color{delim}{]}}}}{1},
}

%% file: tex/abstract.tex
\begin{abstract}

In recent years, the rapid advancement of deep neural networks (DNNs) has revolutionized artificial intelligence, enabling models with unprecedented capabilities in understanding, generating, and processing complex data. These powerful architectures have transformed a wide range of downstream applications, tackling tasks beyond human reach. In this paper, we introduce \textbf{Omniwise}, the first end-to-end, self-supervised fine-tuning pipeline that applies large language models (LLMs) to GPU kernel performance prediction—a novel use case in performance profiling. Omniwise is model-agnostic and lightweight, achieving strong results even with a small 3B-parameter model. It can predict key performance metrics, including memory bandwidth, cache hit rates, GFLOPs, and arithmetic intensity, directly from kernel code without the need for code execution or profiling tools. Our approach achieves over 90\% of predictions within 10\% relative error on GPU kernels executed on AMD MI250 and MI300X architectures. In addition to the pipeline, we develop an online inference server and a Visual Studio Code plugin that seamlessly integrate LLM-based performance prediction into developers' workflows.

\end{abstract}

%% file: tex/intro.tex
\section{Introduction}
\label{sec:intro}

As the scale and complexity of modern computation continue to increase, optimizing the performance of individual kernel codes may yield limited overall improvements due to bottlenecks arising from various emerging graphical processing units (GPUs) and hardware systems on a chip (SoCs) novel architectures, compiler options, and other factors influencing the performance of a GPU kernel. Traditional approaches often rely on the experience and intuition of developers, which, while valuable, cannot easily capture the diverse and abstract computation motifs found in advanced workloads. Additionally, expert developers typically lag behind the rapidly changing machine learning (ML) algorithms and code developed by ML experts that need optimization \cite{reddi2020mlperf, wang2024preliminary}. These challenges become particularly pressing in scenarios that require correlating a complex set of labels—such as performance counters and environment configurations—with a single input code snippet.

\paragraph{GPU Kernels Performance Analysis and Optimizations.} Within the realm of high-performance computing, GPU kernel optimization is critical but notoriously difficult. Vendors provide tools such as ROCm Compute Profiler~\cite{Lu:2024:RRC} and Nsight Compute~\cite{NVIDIA:2024:NNC} to aid developers; however, these tools require complete code development, debugging, and repeated kernel execution for profiling. Moreover, profiling and optimizing a GPU kernel requires a deep understanding of the GPU microarchitecture, an additional expertise that not many developers have -- combining that with new GPU microarchitecture releases further limits the ability of profiling to fewer ``ninja'' developers. Since collecting performance counters involves replaying kernels (or entire applications) multiple times, profiling can consume hours or even days for large-scale problems. In AMD's ROCm Compute Profiler~\cite{Lu:2024:RRC}, collecting all counters can require replaying the application approximately 14 times, thus requiring at least 14x the application runtime and the profiling overhead. Moreover, AMD's profilers rely on application-level replay rather than kernel-level replay, making the process even more expensive.

\paragraph{Our Approach: \omniwise.} In response to these limitations, we propose \omniwise, a novel, end-to-end pipeline that leverages large language models (LLMs) to identify, represent, and predict high-level computation motifs. By harnessing LLMs' pattern-recognition capabilities, \omniwise is able to predict GPU code's performance counters at real-time, dramatically reducing the clock-time required to collect performance counters. This enables developers to rapidly identify and respond to performance degradation in their code, instead of waiting hours. Unlike traditional performance profiling methods, \omniwise doesn't require complete code development and debugging before prediction  as the intention aims for capturing both high-level and low-level computational motifs. Furthermore, \omniwise is able to achieve 90\% of predicted counters falling within 10\% of relative error compared to the ground-truth in our test set, making it a practical and efficient alternative to traditional profiling workflows for GPU software developers. In this paper, we will reveal \omniwise, our end-to-end pipeline, starting from data collection in Section \ref{ssec_data_augmentation}, to fine tuning techniques in Section \ref{ssec_model_training}, then to our results in Section \ref{sec:results}.

Our contributions are as follows:

\begin{enumerate}
    \item Novel and end-to-end pipeline and methodology for LLM-powered profiling tool, 
    \item A fine-tuned \llama 3.2 3B model for predicting performance counters with 90\% accuracy, 
    \item Extensive evaluation of the model's performance in predicting HIP kernel's performance. 
    \item A Visual Studio code plugin for serving the model to developers. 
\end{enumerate}

%% file: tex/background.tex
\section{Background and Previous Work}
\label{sec:background}

This section will first provide a high-level overview of the GPU microarchitecture details and GPU performance counters. Then, we will cover the role of large language models (LLMs) in performance analysis tasks and model choices. 

\subsection{GPUs and Performance Counters}
\subsubsection{GPU Architecture}
Conceptually, we can divide a GPU architecture into compute and memory hierarchies. Although various architectures slightly differ in their description, we focus on the MI200 GPU series for simplicity, but the description mostly applies to other GPUs. Readers can refer to the documentation and white papers for per-GPU specification~\cite{ROCm:2025:PMD}. Typically, in modern GPUs, the memory hierarchy starts with registers and L1 cache, which are a common resource for a compute unit. Registers are local to a thread, while threads share the L1 cache on a compute unit. The L1 cache is typically accompanied by a software-managed cache (i.e., Local data share or LDS) at similar speeds and shared across a workgroup (i.e., block). The L2 cache is shared across all compute units and is typically considered a \emph{coherence point} since it is visible to all threads on the GPU device. A last-level cache may also be present in more modern GPUs, but in this paper, we only focus on traditional memory hierarchies. Finally, main memory is the last level of the memory hierarchy, accessible to the entire GPU device. For the compute hierarchy, each compute unit (CU) is divided into SIMD units (e.g., 4 for MI250) that process 16-wide SIMD instructions. Each CU contains Vector arithmetic logic units (VALU) that process the SIMD instructions. Additionally, CUs contain a register file and vector memory units as well as other such units for scalar arithmetic logic units (SALU) and specialized hardware such as matrix cores. 

\subsubsection{Performance Counters}
To quantify performance, \omniwise borrows ROCm Compute Profiler's~\cite{Lu:2024:RRC} empirical roofline model and derived performance metrics. In contrast to theoretical rooflines that use the theoretical achievable peak performance, an empirical roofline uses micro-kernels to measure the \emph{attainable} peak performance on the hardware. The roofline performance model relates floating-point performance (GFLOP/s) as a function of machine peak performance, machine peak bandwidth, and arithmetic intensity of the application kernels. Arithmetic intensity can be defined as the ratio of total floating-point operations (FLOPs) performed by a given kernel in VALUs, to the total data movement (Bytes) required to support those FLOPs~\cite{Yang:2018:AER}. Each kernel is represented as a point on the coordinate plane. We find the empirical roofline to be an effective medium for communicating application performance because of its ability to relate measured kernel performance to the \emph{attainable} peak performance of target hardware. 

The \omniwise model also presents derived metrics such as cache hit rates which can be helpful in understanding the number of cache line requests that are hitting relative to the total number of incoming cache line requests.

\omniwise avoids the tedious process of profiling code and collecting hardware counters on actual hardware by encoding the complex set of code labels including the system description (e.g., ROCm versions, compiler flags, GPU architecture, etc.). Tracking these configuration settings is integral to predicting the high-level counters with LLMs\@.

\subsection{Large Language Models}

LLMs such as OpenAI's ChatGPT 4~\cite{Achiam:2023:GTR}, Google's Gemini 1.5~\cite{Team:2024:GUU}, and Meta's \llama 3~\cite{Dubey:2024:L3H} have demonstrated exceptional capabilities in recognizing patterns across diverse domains and complicated scenarios. By leveraging these model's ability to capture complex relationships, we can address challenges like correlating performance counters with source code across varying hardware and software configurations. 

Because LLMs exist in different flavors, model choice in our case-specific scenario becomes a vital decision. We explored the merits of general-purpose models (e.g., \llama) that are trained on generic datasets against special-purpose models for specific domains such as StarCoder for code-related tasks. Our hypothesis is that general purpose LLMs when fine-tuned using our \omniwise pipeline will outperform special purpose models. We will next highlight the important characteristics of general- and special-purpose LLMs and how each plays a different role in predicting GPU performance counters.

\subsubsection{General-Purpose LLMs}
General-purpose LLMs, such as ChatGPT~\cite{Team:2024:GUU}, Gemini~\cite{Team:2024:GUU}, Qwen~\cite{Yang:2024:QTT}, \llama~\cite{Dubey:2024:L3H}, etc., are designed to excel across a wide variety of tasks, and are trained on a vast dataset. For our pipeline, we prioritize models with open-source checkpoints to ensure data security, ease of customization, and the accessibility and downstream task performance. Among the available options, we selected \llama for its strong performance in natural language, coding ability, and ease of use. In particular, we use the \llama 3.2 3B Instruct model~\cite{HF:2024:L32} because it is lightweight for deployment and offers a large 128K-token context window, striking a balance between performance and portability. Its compact size also makes it feasible to serve the model on modern laptops.

\subsubsection{Special-Purpose LLMs}

In addition to general-purpose models, we explored coding-specific LLMs tailored to our code-centric scenario. These models are pretrained on extensive code-related datasets and are designed to handle programming tasks effectively. Examples include StarCoder-2~\cite{Lozhkov:2024:SSC}, DeepSeekCoder~\cite{Zhu:2024:DBC}, and CodeBERT~\cite{Feng:2020:CAP}. After evaluating their capabilities, we chose StarCoder-2 for its superior coding proficiency and its extensive pretraining on a large volume of code data, which we anticipate enables a deeper intrinsic understanding of performance-related arguments in code.

\subsubsection{Model Choice for Omniwise}
After considerable empirical studies in \llama 3.2 3B and StarCoder2 3B on our downstream tasks, \llama 3.2 3B has exceeded StarCoder2 3B in the accuracy of predictions as well as the ability of following instructional commands. We attribute this to the limited amount of natural language corpus (mostly GitHub issues and comments) and lack of reinforcement learning (RL) trained on StarCoder2. Moreover, \llama 3.2 3B has a larger context window of 128K (StarCoder context window is only 16K). As a result, we will be showing our fine-tuned results on \llama 3.2 3B Instruct in the rest of the papers.

\subsubsection{Fine-tuning LLMs}
\label{ssec_fine_tuning_llms}
Fine-tuning techniques, such as Low-Rank Adaptation (LoRA)~\cite{Hu:2021:LLR}, a parameter-efficient fine-tuning (PEFT) method, enables efficient domain-specific customization of LLMs by reducing the number of tunable parameters without sacrificing core capabilities. The ease of fine-tuning is also attributed to Deepspeed's ZeRO \cite{rasley2020deepspeed} support. This adaptability makes LLMs ideal for powering the Omniwise pipeline, which automates and accelerates GPU kernel performance analysis and optimization.

\subsubsection{Other relevant work.} More recently, Meta released the Meta LLM Compiler~\cite{Cummins:2024:MLM} which is the most similar previous work to \omniwise.  Meta's LLM Compiler correlates LLVM's intermediate representations (IRs) to code size; in their work, their 7B model achieves 0.083 and 0.225 mean absolute percentage error (MAPE) when predicting unoptimized and optimized binary sizes respectively. In contrast to the Meta compiler, we focus on correlating source code (e.g., HIP~\cite{ROCm:2025:HIP}) to multiple performance counters, not only a single one.

%% file: tex/omniwise.tex
\section{\omniwise}
\label{sec:design}

\begin{figure*}
  \centering
  \includegraphics[page=1, trim=5 120 5 110, clip, width=\textwidth]{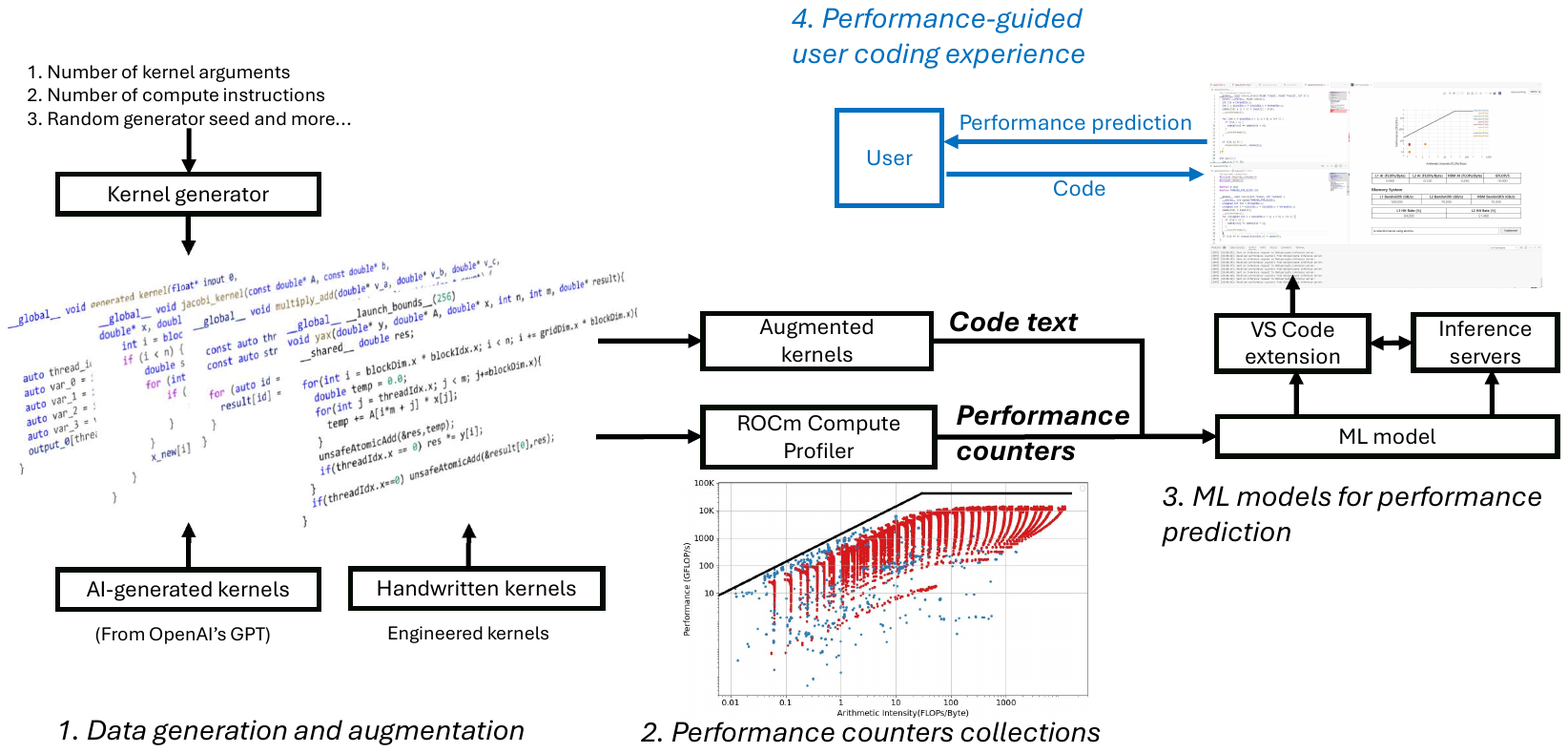}
  \caption{\omniwise end-to-end workflow.}
  \label{fig:end-to-end}
\end{figure*}

\begin{figure*}
    \centering
    \begin{minipage}[t]{0.32\textwidth}
        \centering
        \begin{lstlisting}[caption={Synthetic}, label={lst:synthetic}]
#include <cstdint>
#include <iostream>

#include <hip/hip_runtime.h>
#include <hip/hip_runtime_api.h>
#include <thrust/device_vector.h>

__global__ void generated_kernel(double *input_0, double *output_0) {
 auto thread_id = threadIdx.x + 
                   blockIdx.x * blockDim.x;
 auto var_0 = input_0[thread_id];
 auto var_1 = var_0 * var_0 + var_0;
 output_0[thread_id] = var_1;
}
int main(int, char **) {
 thrust::device_vector<double> input_0(102528, 1);
 thrust::device_vector<double> output_0(102528, 1);
 
 std::size_t block_size{256};
 std::size_t input_size{102528};
 std::size_t num_blocks{(input_size + block_size - 1) / block_size};
  
 generated_kernel<<<num_blocks, block_size>>>(
    input_0.data().get(), output_0.data().get());
    
 auto status = hipDeviceSynchronize();
 if (status != hipSuccess) {
  std::cout << "kernel launch failed\n";
 }
}
        \end{lstlisting}
    \end{minipage}
    \hfill
    \begin{minipage}[t]{0.32\textwidth}
        \centering
        \begin{lstlisting}[caption={AI-Generated}, label={lst:ai_generated}]
#include <hip/hip_runtime.h>
#include <iostream>
#include <vector>

__global__ void reduce_kernel(float* input, float* output, int n) {
 unsigned int i = blockIdx.x * blockDim.x + threadIdx.x;
 if (i < n) {
  atomicAdd(output, input[i]);
 }
}

int main() {
 const int N = 1 << 24;
 const int threadsPerBlock = 1024;
 const int blocks = (N + threadsPerBlock - 1) / threadsPerBlock;
 size_t size = N * sizeof(float);
 std::vector<float> h_input(N, 1.0f);
 float* d_input;
 float* d_output;
 hipMalloc(&d_input, size);
 hipMalloc(&d_output, sizeof(float));
 hipMemcpy(d_input, h_input.data(), size, hipMemcpyHostToDevice);
 hipMemset(d_output, 0, sizeof(float));
 reduce_kernel<<<blocks, threadsPerBlock>>>(d_input, d_output, N);
 float h_output;
 hipMemcpy(&h_output, d_output, sizeof(float), hipMemcpyDeviceToHost);
 std::cout << "Sum: " << h_output << std::endl;
 hipFree(d_input);
 hipFree(d_output);
}
    \end{lstlisting}
    \end{minipage}
    \hfill
    \begin{minipage}[t]{0.32\textwidth}
        \centering
        \begin{lstlisting}[caption={Hand-written}, label={lst:hand_written}]
#include <thrust/device_vector.h>
#include <cstdint>
#include <iostream>

__global__ void add(double* v_a, double* v_b, double* result,
                    std::size_t count) {
 const auto thread_id = threadIdx.x + blockIdx.x * blockDim.x;
 const auto stride = blockDim.x *
                     gridDim.x;
 for (auto id = thread_id; id < count;
           id += stride) {
  result[id] = v_a[id] + v_b[id];
 }
}

int main() {
 const std::size_t count{1'000'000};
 thrust::device_vector<double> v_a(count,
                                   1);
 thrust::device_vector<double> v_b(count,
                                   3);
 thrust::device_vector<double> result(
                                count);
 const std::size_t block_size{512};
 const std::size_t num_blocks{1024};
 add<<<num_blocks, block_size>>>(
 v_a.data().get(), v_b.data().get(), result.data().get(), count);
 const auto status = hipDeviceSynchronize();
 if (status != hipSuccess || result[0] != 4 || result[count - 1] != 4) {
  std::cout << "Kernel failed.\n";
  return -1;} else {
  std::cout << "Success!\n";}}
    \end{lstlisting}
    \end{minipage}
    \caption{\omniwise dataset examples.}
    \label{fig:code_examples}
\end{figure*}

In this paper, we introduce \omniwise, an end-to-end workflow that leverages fine-tuned LLMs to provide developer's GPU code with interactive performance counters. ~\figref{end-to-end} shows the \omniwise workflow. In it, we see the main four stages of \omniwise:
\begin{inparaenum}[(1)]
    \item data generation and augmentation, 
    \item performance counters collection, 
    \item model training, and
    \item model serving.
\end{inparaenum} In this section, we will discuss the various components and how they interact.

\subsection{Data Generation and Augmentation}
\label{ssec_data_augmentation}

To enable accurate performance prediction of GPU kernels, \omniwise\ requires a dataset that reliably associates code with corresponding hardware performance counters. We constructed a dataset of approximately 955,000 HIP kernels through a combination of data collection and targeted augmentation. The dataset was split into 90\% for training and 10\% for validation, with around 4000 kernels reserved for testing.

While numerous open-source GPU codebases exist, they are often not suited for our use case due to their reliance on external libraries, fragmented code structure, or lack of measurable performance ground-truth. Instead, we curated a collection of self-contained kernels, each embedded in a single source file to ensure clean parsing and to simplify the LLM source-to-counters prediction task. For more complex applications, similar datasets can be constructed by extracting flattened kernel code via compiler preprocessing outputs.

\paragraph{Challenges of Scale and Label Correlation.} Despite the promise of LLMs, constructing the necessary datasets for fine-tuning presents two critical challenges. \textbf{First}, scale poses a major hurdle: to train a model that can comprehensively predict counters for various architectures and computational motifs, one would need a dataset that captures the full range of code and counter variations. \textbf{Second}, label correlation is non-trivial: each code snippet must be associated with a potentially large set of performance counters, environmental variables, and hardware descriptors. Existing solutions rarely tackle the problem of mapping a complex label set onto a single input code snippet while also capturing inter- and intra-code relationships across multiple inputs.

To address these challenges, \omniwise takes a three-pronged approach to dataset generation: 

\paragraph{Synthetic Kernels.} We build a simple synthetic kernel generator that generates GPU code given the number of kernel arguments, dataset size, data types, and the number of memory and compute instructions. To generate the GPU kernel, the kernel generator randomly selects kernel load and store operations from and to the kernel arguments, then samples compute instructions and their operands from nearby lines. Sampling is performed in a skewed way towards recent variables to ensure a data-dependency chain. \lsref{synthetic} shows an example of a generated kernel using the synthetic kernel generator.

\paragraph{AI-Generated Code.} We leverage OpenAI's GPT-4o~\cite{OpenAI:2024:HGO} to build a collection of GPU kernels, including their host-side code that drives the execution. We use the following system prompt:
\begin{quote}
\emph{
You are a skilled GPU programmer. Your response will be a GPU kernel with the proper includes and main file. The file should compile and run and use double or float data types and large problem sizes. Your response should not contain any special markdown formatting like ```cpp or comments.
}
\end{quote}

Using the system prompt and OpenAI's Python APIs, we build the AI-generated portion of the dataset using a prepared list of problems such as reductions, scans, and General Matrix Multiply (GEMM) while varying the temperature for answer generation alongside instructing the model to generate solutions at various optimizations levels. For example, sample prompts may look like:
\begin{itemize}
    \item Generate a simple reduction kernel
    \item Generate a reduction kernel that uses shared memory
    \item Generate a highly-optimized reduction kernel
\end{itemize}
In our experiments, only 7\% of the generated code did not successfully compile or failed while profiling -- these kernels were simply excluded from the dataset. Note that we are interested in correlating performance counters to source code patterns, so we do not attempt to validate the code. Moreover, code written by humans may also contain bugs. That being said, we observed high-quality kernels when we manually inspected the dataset. \lsref{ai_generated} shows a simple AI-generated reduction kernel. 

\paragraph{Custom Kernels.} To further enrich our dataset, we use examples from AMD's training examples~\cite{AMD:2025:HTE} as well as custom hand-written kernels that explore different GPU programming paradigms (e.g., per-thread assignment per-thread processing, per-thread assignment per-tile processing~\cite{Awad:2023:AAI}, and persistent kernels~\cite{Gupta:2012:ASO}) and common optimization strategies (e.g., privatization techniques).

Additionally, \omniwise leverages the combinatorial explosion that arises when profiling the same kernel on various architectures when compiled with various optimization
flags. For the same kernel, we profile it after compiling the code with compilation flags such as \texttt{-ffast-math} and \texttt{-munsafe-fp-atomics}, and for two different architectures (MI250 and MI300X). We further enlarge the dataset and make it variable-name agnostic by replacing variable names with randomly generated ones. Furthermore, when mapping the counters to the source code, we adopt a JSON-based formatting way when attaching performance counters to the source code (see \secref{data_cleaning}). \figref{code_examples} shows an example of the dataset.

Note that across these datasets, we maintained a large problem size, and additionally, for simplicity, we maintained a consistent datatype whenever possible. Particularly for the large problem size, the rationale is that, typically, programmers offload large enough problems to utilize the GPU fully -- that being said, we can extend the same workflow beyond our assumptions here, but we leave these for future exploration. We will discuss what our assumptions mean for users in \secref{conclusion}.

\subsection{Performance counters collection}

Once we have the kernels, we compile and profile each sample in the dataset. Beside the code itself, performance of a GPU kernel is a function of many other parameters such as toolkit versions (e.g., ROCm's version), compiler versions, compiler options and hardware. These various parameters affecting the performance enables us to turn each sample into multiple by compiling using various combinations of compiler flags, as well as profiling the code on various GPU architecture. 

\begin{table}[ht]
    \centering
    \begin{tabular}{@{} l c @{}}
        \toprule
        \textbf{Specification} & \textbf{\llama 3.2 3B Instruct} \\
        \midrule
        \textbf{Model Size} & 3 billion parameters \\
        \textbf{Context Length} & 128,000 tokens \\
        \textbf{Vocabulary Size} & 128,000 tokens \\
        \textbf{Number of Layers} & 32 \\
        \textbf{Number of Attention Heads} & 24 \\
        \bottomrule
    \end{tabular}
    \caption{Key Architectural Specifications of \llama 3.2 3B Instruct.}
    \label{tab:llama32_3b_specs}
\end{table}

\begin{table}[ht]
    \centering
    \caption{Performance metric ranges before normalization.}
    \begin{tabular}{l@{} l c @{}}
        \toprule
         &\textbf{Metric} & \textbf{Range}\\
        \midrule
         &\text{L1 Cache Hit Rate (\%)}  & 0 -100\\
         &\text{L2 Cache Hit Rate (\%)}  & 0 - 100\\
         \midrule
         &\text{L1 Cache Bandwidth (GB/s)}  & 0 - 16384\\
         &\text{L2 Cache Bandwidth (GB/s)}  & 0 - 16384\\
         &\text{HBM Read Bandwidth (GB/s)}  & 0 - 16384\\
         &\text{HBM Write Bandwidth (GB/s)}  & 0 - 16384\\
         \midrule
         &\text{L1 Arithmetic Intensity (FLOPs/Byte)}  & 0 - 2048\\
         &\text{L2 Arithmetic Intensity (FLOPs/Byte)}  & 0 - 5120\\
         &\text{HBM Arithmetic Intensity (FLOPs/Byte)}  & 0 - 2048\\
         \midrule
         &\text{L1 GFLOP/S}  & 0 - 12288\\
         &\text{L2 GFLOP/S}  & 0 - 12288\\
         &\text{HBM GFLOP/S}  & 0 - 12288\\
        \bottomrule
    \end{tabular}
    \label{tab:norm_ranges}
\end{table}

\subsection{Data Normalization and Formatting}
\label{sec:data_cleaning}
\paragraph{Data Normalization} To enhance model interpretability across diverse and complex performance metrics, we apply a standardized pre-processing step in which all collected metrics are normalized. Details can be found in Table~\ref{tab:norm_ranges}. We convert raw numerical values into a normalized format within the range of `0.xxx' relative to their respective hardware ceilings. This mitigates issues where the model might otherwise predict values near 1.0 or 0.0, which could lead to order-of-magnitude errors due to the exponential nature of these quantities. Some units are implicitly defined relative to the hardware's theoretical peak values. As an example, the L1 cache hit rate is naturally bounded at 1.00, corresponding to a 100\% hit rate. 

\paragraph{Data Formatting} To ensure output stability and consistency in downstream tasks—regardless of decoding parameters such as temperature, top-k, or nucleus sampling—we constrain the model's output format. Through empirical evaluation of multiple prompting strategies for encoding performance counters, we found that enforcing a strict JSON schema yields the most reliable outputs. While this approach may incur a slight degradation in predictive accuracy compared to more flexible prompting formats, it substantially reduces failure cases in downstream VS-Code serving infrastructure. The JSON-constrained output significantly improves format consistency across varied decoding configurations, demonstrating its robustness even under stochastic generation settings. \figref{training_sample} shows an example of a training dataset sample where we combine the system and user prompts, code, compiler flags, and performance counters formatted in JSON.

\begin{figure}[h!]
    \centering
    \begin{minipage}{0.8\textwidth}
        \begin{verbatim}
<|begin_of_text|><|start_header_id|>system
<|end_header_id|>You are an expert GPU programmer
and profiler. Given a code, you will predict its
performance counters.<|eot_id|>
<|start_header_id|>user<|end_header_id|>
For the GPU architecture gfx90a and the compiler flags
--std=c++17 -O3 -ffast-math, what are the
bandwidth, arithmetic intensity, hit rates, flops
of the following code? Output the answer in JSON format.

% Code removed for brevity

<|eot_id|><|start_header_id|>assistant<|end_header_id|>
Here are the performance counters for the gfx90a GPU
architecture and the --std=c++17 -O3 -ffast-math
compiler flags combination in JSON format: 

```json
{

"compiler_flags": "--std=c++17 -O3 -ffast-math",
"architecture": "gfx90a",
"L1_Cache_Arithmetic_Intensity": "0.002",
"L2_Cache_Arithmetic_Intensity": "0.002",
"HBM_Arithmetic_Intensity": "0.004",
"L1_Cache_GFLOPS": "0.459",
"L2_Cache_GFLOPS": "0.459",
"HBM_GFLOPS": "0.459",
"L1_Cache_Bandwidth": "0.089",
"L2_Cache_Bandwidth": "0.070",
"L2_Fabric_Write_BW": "0.022",
"L2_Fabric_Read_BW": "0.022",
"L1_Cache_Hit_Rate": "0.500",
"L2_Cache_Hit_Rate": "0.370"
}
```<|eot_id|>
    \end{verbatim}
    \end{minipage}
    \caption{\omniwise training sample.}
    \label{fig:training_sample}
\end{figure}

\subsection{Model training}
\label{ssec_model_training}

Given the architected dataset, we now have a suitable input for fine-tuning. 
We use the \llama 3.2 model~\cite{Meta:2024:L32} and we specifically pick the \emph{instruct} 3 billion parameters model. Although various models can be used as a starting point, the open-source \llama 3.2 with only 3 billion parameters and 128K context length provides enough power for \omniwise. We use the instruct model since it better serves our goal of JSON-formatted output. A summary of the model architecture is shown in Table~\ref{tab:llama32_3b_specs}.  Next, we tokenize the dataset then fine-tune the \llama model.

\paragraph{Training methodology.} All the fine-tuning workload has been on 1 compute node with 4 MI250s GPUs, each contains 2 Graphic Compute Dies (GCDs)\@, and each GCD contains 64GB of HBM. We deployed Deepspeed ZeRO \cite{Rajbhandari:2020:ZMO} to prevent GPU out of memory, while maintaining a high system throughput. We adapt AdamW \cite{adam} and cosine learning rate scheduler with 10\% of warm-up steps. Details of the hyperparameters can be found in Table~\ref{tab:training_params}.

\begin{table}[ht]
    \centering
    \caption{Hyperparameters used to fine-tune \llama 3.2 3B Instruct.}
    \begin{tabular}{l@{} l c @{}}
        \toprule
         &\textbf{Training Parameter} & \textbf{Value}\\
        \midrule
         &\text{Epochs}  & 1\\
         &\text{Max steps}  & 50,000\\
         &\text{Learning rate}  & 5e-5 \\
         &\text{Batchsize per device}  & 2 \\
         &\text{BF16}  & true \\
         &\text{Random seed}  & 42 \\
         &\text{LoRA's r dimension}  & 24 \\
         &\text{beta\_1}  & 0.9 \\
         &\text{beta\_2}  & 0.999 \\
        \bottomrule
    \end{tabular}
    \label{tab:training_params}
\end{table}

\subsection{Serving}

We implemented a Visual Studio Code (VSCode) extension to serve as the final product through which we will serve the model. The extension is bundled with an inference server that performs the predictions using the fine-tuned \omniwise model. The specifics of the VSCode extension are irrelevant to this paper; however, at a high-level, the extension listens to text edit events and on each change, it sends an inference request to the inference server. The inference request includes the source code and extension user-controlled variables such as compiler flags and architecture. Upon receiving an inference request, the inference server simply performs a forward pass, extracts the JSON-formatted performance counters, denormalizes the counters and finally returns the predictions to the extension. The extension provides the interactive predictions via a roofline plot as well as tabular data providing high-level performance predictions of memory and compute subsystems to guide the developers as they write and modify their code.

%% file: tex/results.tex
\section{Results}
\label{sec:results}
Here we are representing our preliminary results of running on finetuned \llama 3.2B on our test set.  We report relative error-based accuracy and discuss model behavior across memory and compute dimensions.

\subsection{Experimental Setup}
\label{ssec:exp_setup}

The test set was composed of around 4,000 kernels sampled from the entire pool of data described in Section \ref{ssec_data_augmentation}, and no overlapping with the training set. All reported metrics are normalized relative to peaks shown in~\tabref{norm_ranges}, ensuring that predictions are unitless, comparable, and generalizable.

Predictions are compared against ground-truth performance counters collected using ROCm Compute Profiler, and are normalized relative to the values described in Table~\ref{tab:norm_ranges}, ensuring that predictions are unitless. Evaluation uses the relative error metric:
\[
\text{Relative Error} = \frac{|\text{Predicted} - \text{Ground Truth}|}{\text{Ground Truth} }
\]
We report the proportion of predictions falling below fixed relative error thresholds (2\%, 4\%, 6\%, 8\%, 10\%).

We note that the relative error metric can become numerically unstable when the ground-truth value is close to zero. In such cases, even small absolute deviations may result in disproportionately large relative errors. We observed this effect in a subset of low-range metrics, where a small number of predictions exhibited unusually high relative error despite being qualitatively reasonable. Still, our results show a consistency of high accuracy throughout all the metrics.

\begin{table}[h!]
    \centering
    \caption{Cache Performance and Bandwidth Metrics For Different Errors Percentiles.}
    \label{tab:cache_metrics}
    \begin{tabular}{lccccc}
        \toprule
        \textbf{Metric}            & \textbf{2\%} & \textbf{4\%} & \textbf{6\%} & \textbf{8\%} & \textbf{10\%} \\
        \midrule
        L1 Cache Hit Rate          &              89.9&              91.7&              92.9&              93.6&              94.1\\
        L2 Cache Hit Rate          &              87.3&              91.5&              92.8&              93.4&              94.0\\
        \midrule
        L1 Cache Bandwidth         &              73.8&              82.3&              87.1&              89.5&              90.8\\
        L2 Cache Bandwidth         &              73.5&              82.5&              87.3&              90.3&              91.7\\
        HBM Read Bandwidth         &              76.5&              85.0&              89.5&              91.5&              92.9\\
        HBM Write Bandwidth        &              81.6&              85.3&              90.8&              92.9&              93.8\\
        \midrule
        L1 Arithmetic Intensity    &              99.8&              99.8&              99.8&              99.8&              99.8\\
        L2 Arithmetic Intensity    &              99.4&              99.4&              99.4&              99.4&              99.4\\
        HBM Arithmetic Intensity   &              98.2&              98.2&              98.2&              98.2&              98.2\\
        \midrule
        HBM GFLOP/s                &              84.5&              87.5&              89.8&              91.0&              91.7\\
        \bottomrule
    \end{tabular}
\end{table}

\subsection{Prediction Accuracy Across Metrics}
In this section, we evaluate \omniwise\ along three key dimensions:

\begin{enumerate}
    \item \textbf{Accuracy across diverse performance metrics.} We assess the model's predictive fidelity on ten normalized hardware counters—including cache hit rates, bandwidths, arithmetic intensities, and GFLOPs—collected from diverse kernels and different flags.

    \item \textbf{Robustness to data imbalance and metric sparsity.} We analyze how well the model generalizes across highly skewed and low-range metric distributions, which frequently occur in GPU profiling workloads.

    \item \textbf{Practical viability for profiling-free performance estimation.} We demonstrate that the model achieves high accuracy without relying on runtime execution or instrumentation, enabling real-time performance feedback in developer workflows.
\end{enumerate}

\begin{figure*}[p]
    \centering
    \begin{subfigure}[t]{0.49\textwidth}
        \centering
        \includegraphics[width=1\linewidth]{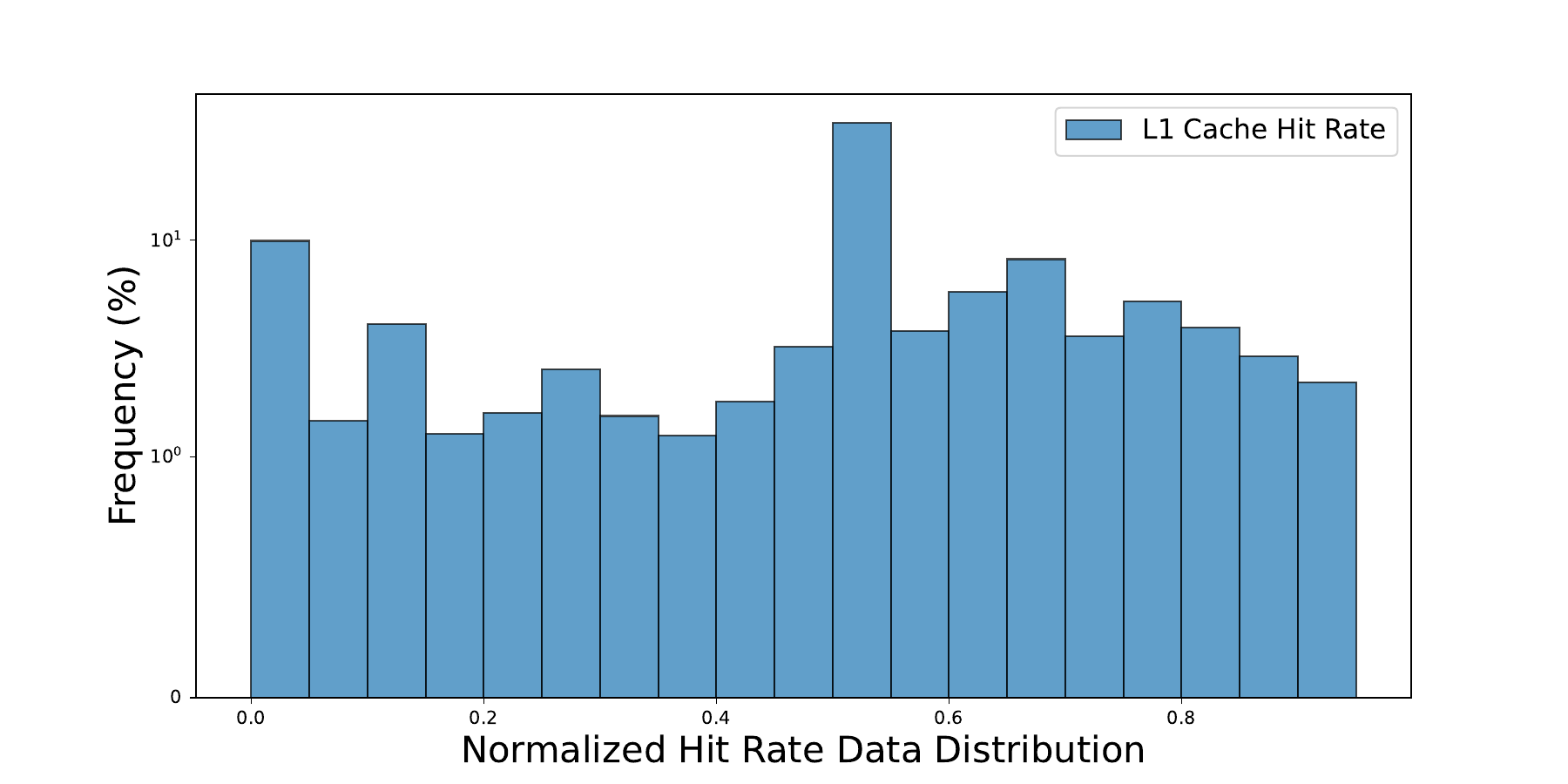}
        \caption{L1 Cache Hit Rate - Input distribution}
    \end{subfigure}
    \begin{subfigure}[t]{0.49\textwidth}
        \centering
        \includegraphics[width=1\linewidth]{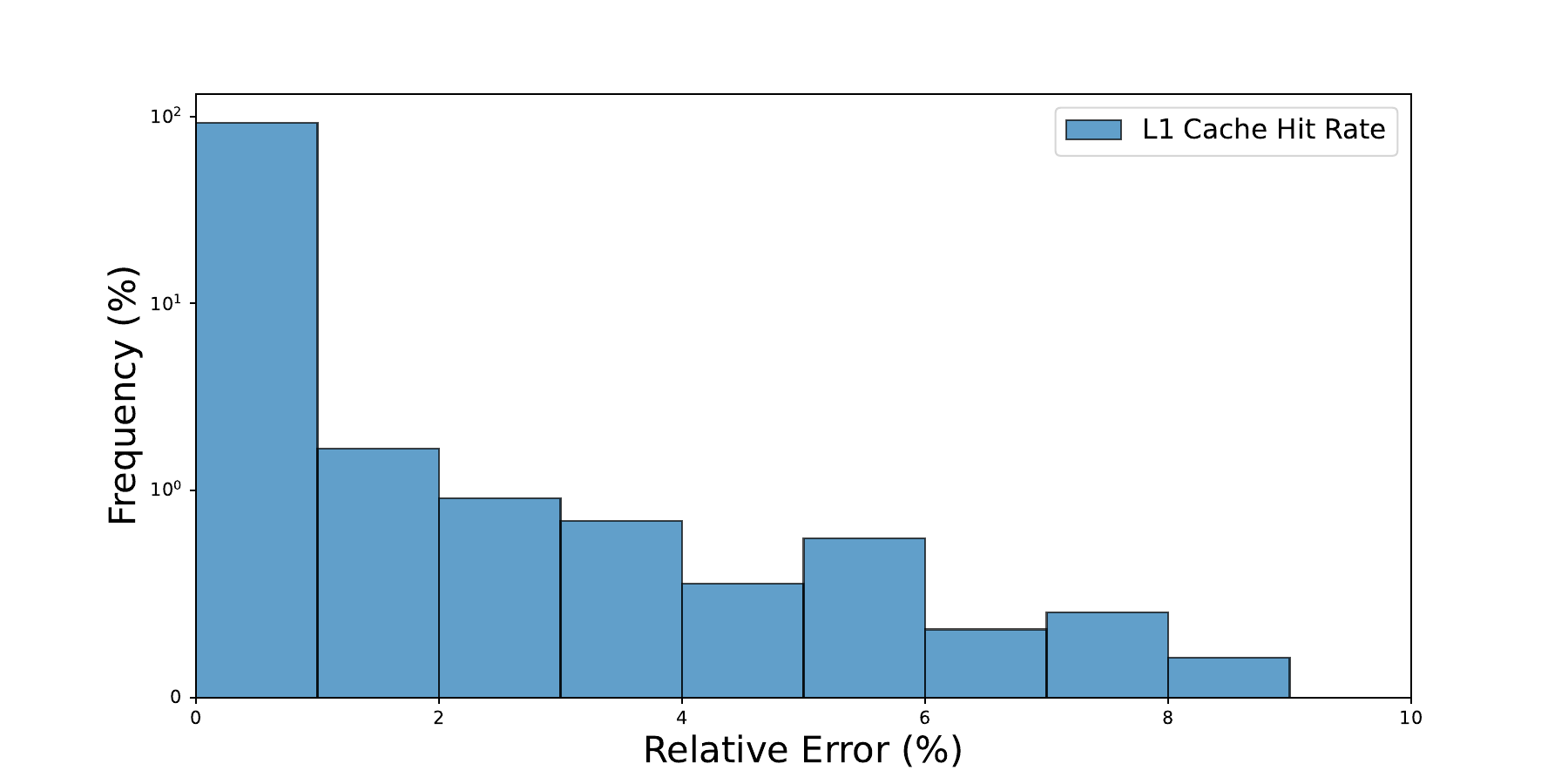}
        \caption{L1 Cache Hit Rate - Relative error}
    \end{subfigure}
    \begin{subfigure}[t]{0.49\textwidth}
        \centering
        \includegraphics[width=1\linewidth]{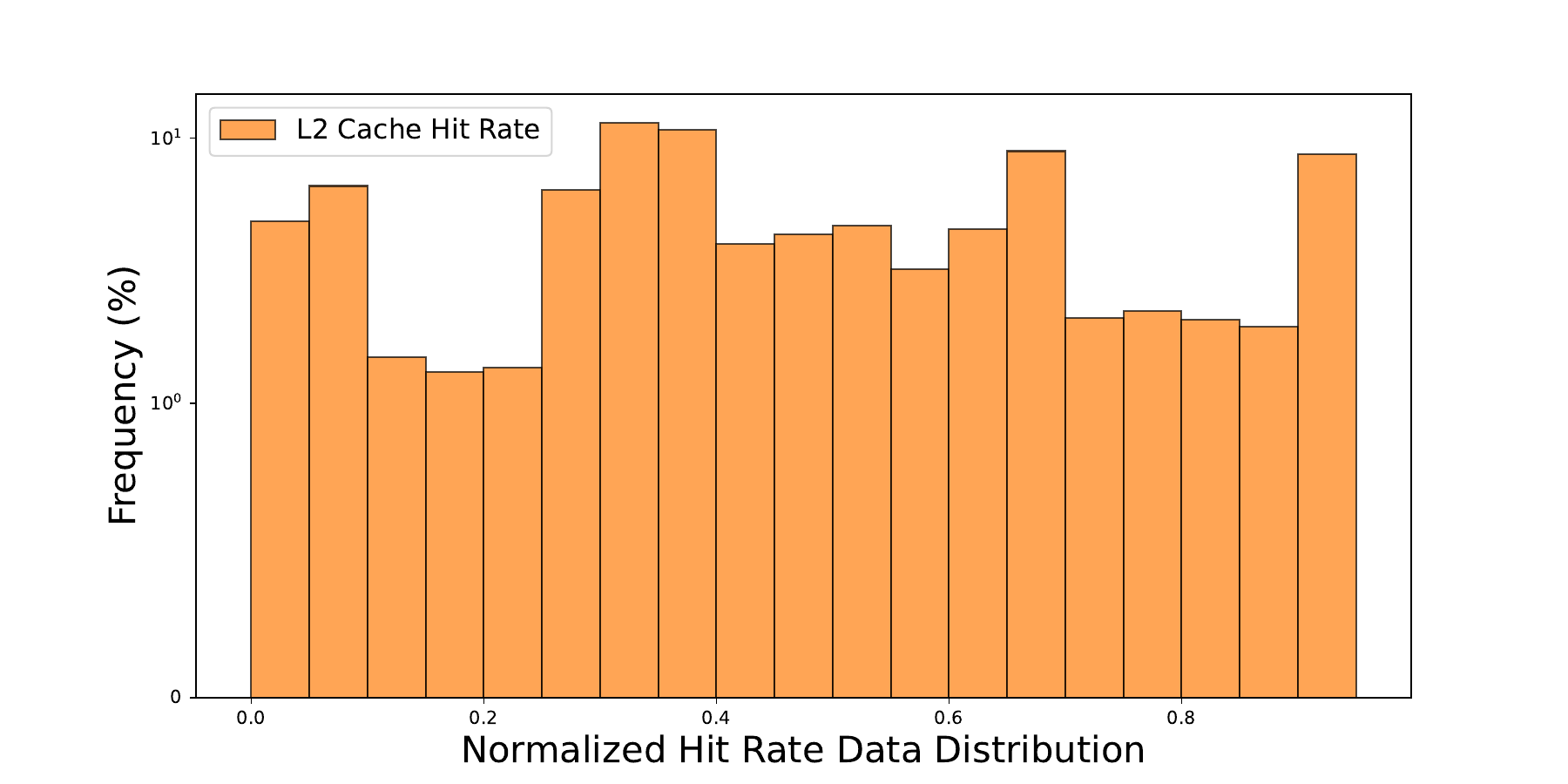}
        \caption{L2 Cache Hit Rate - Input distribution}
    \end{subfigure}
    \begin{subfigure}[t]{0.49\textwidth}
        \centering
        \includegraphics[width=1\linewidth]{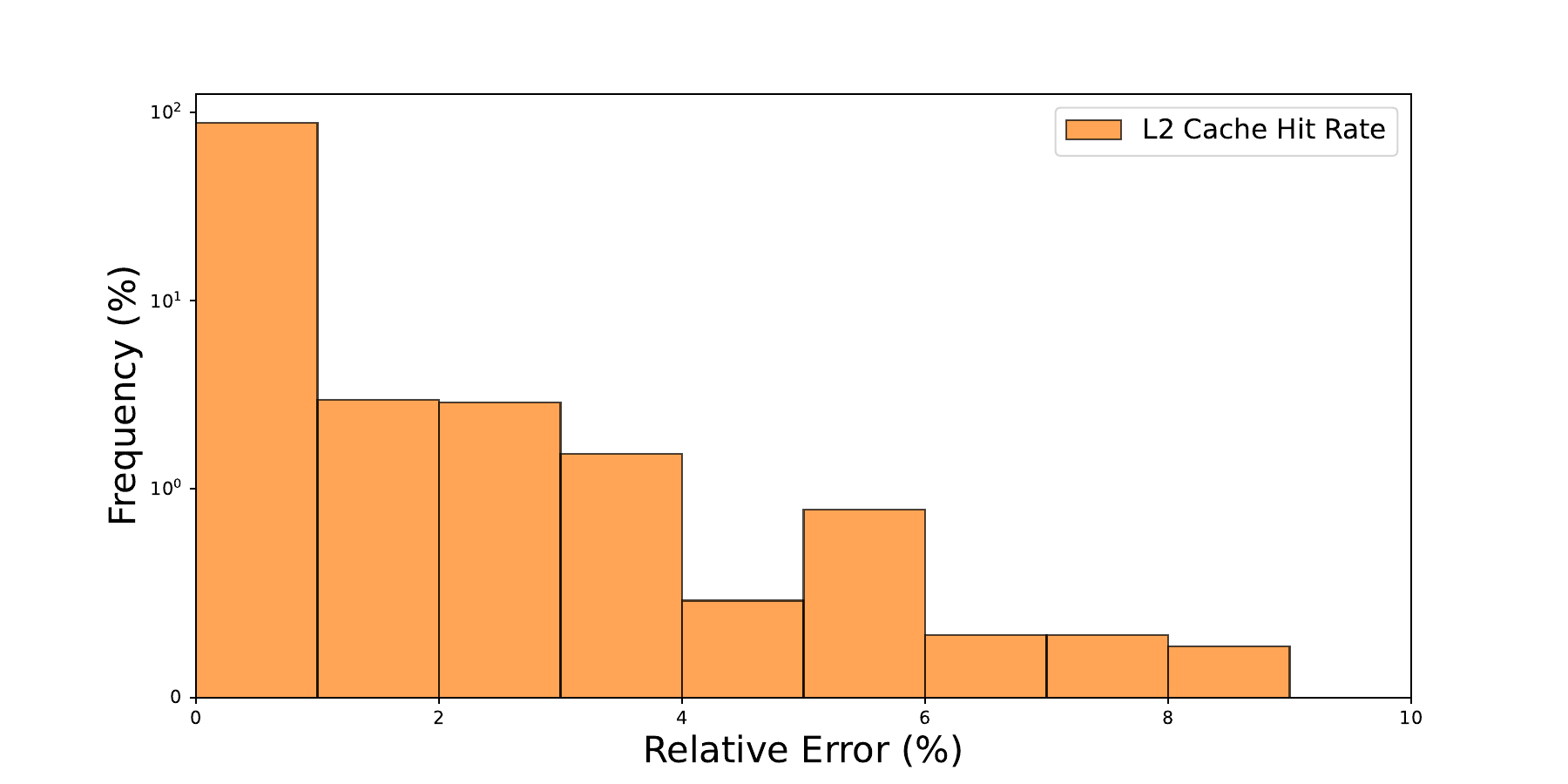}
        \caption{L2 Cache Hit Rate - Relative error}
    \end{subfigure}
    \caption{Distributions and relative errors for L1 and L2 Cache Hit Rates.}
    \label{fig:cache_hit_combined}
\end{figure*}

\begin{figure*}[p]
    \centering
    \begin{subfigure}[t]{0.49\textwidth}
        \centering
        \includegraphics[width=1\linewidth]{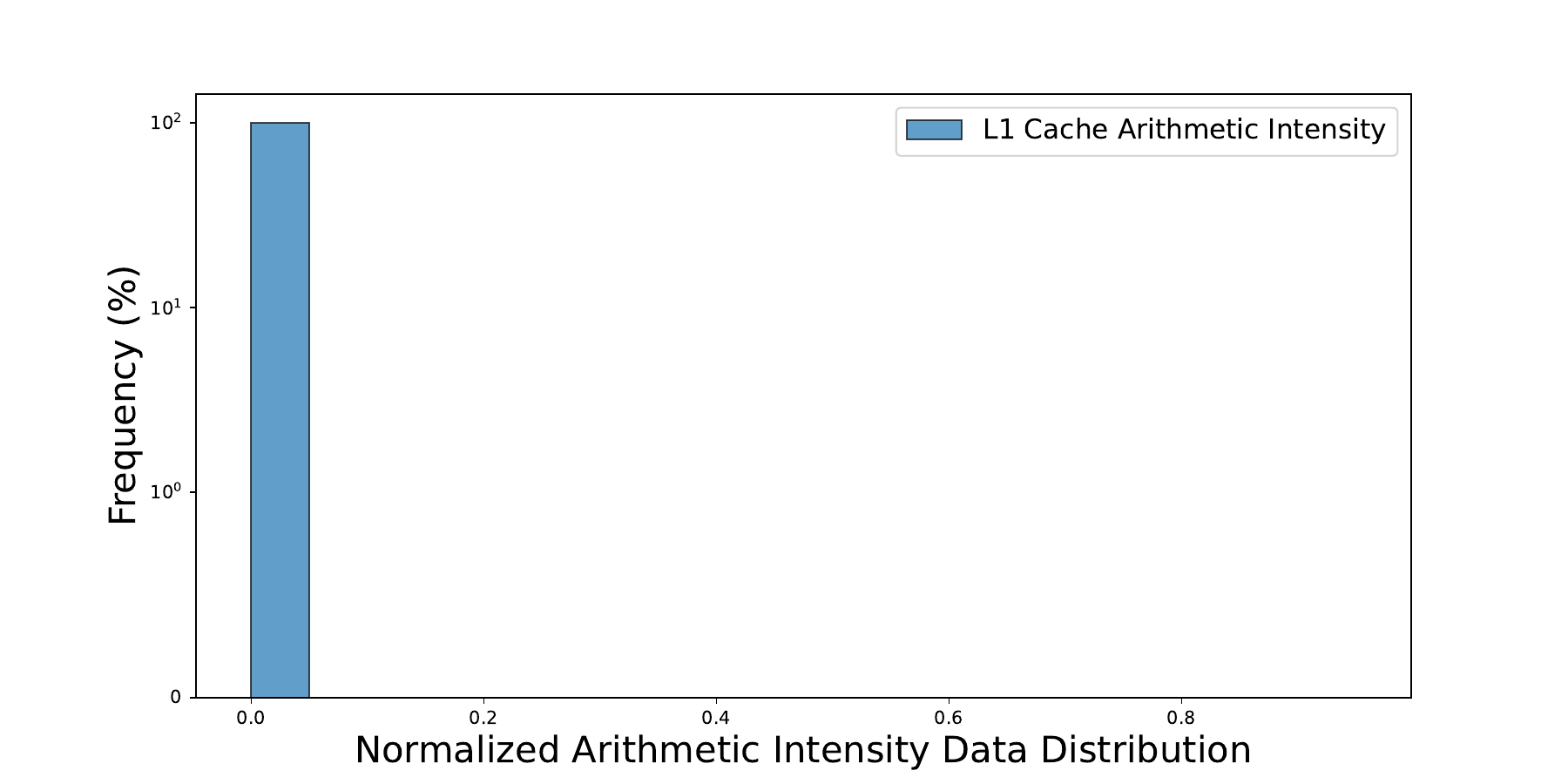}
        \caption{L1 Arithmetic Intensity - Input distribution}
    \end{subfigure}
    \begin{subfigure}[t]{0.49\textwidth}
        \centering
        \includegraphics[width=1\linewidth]{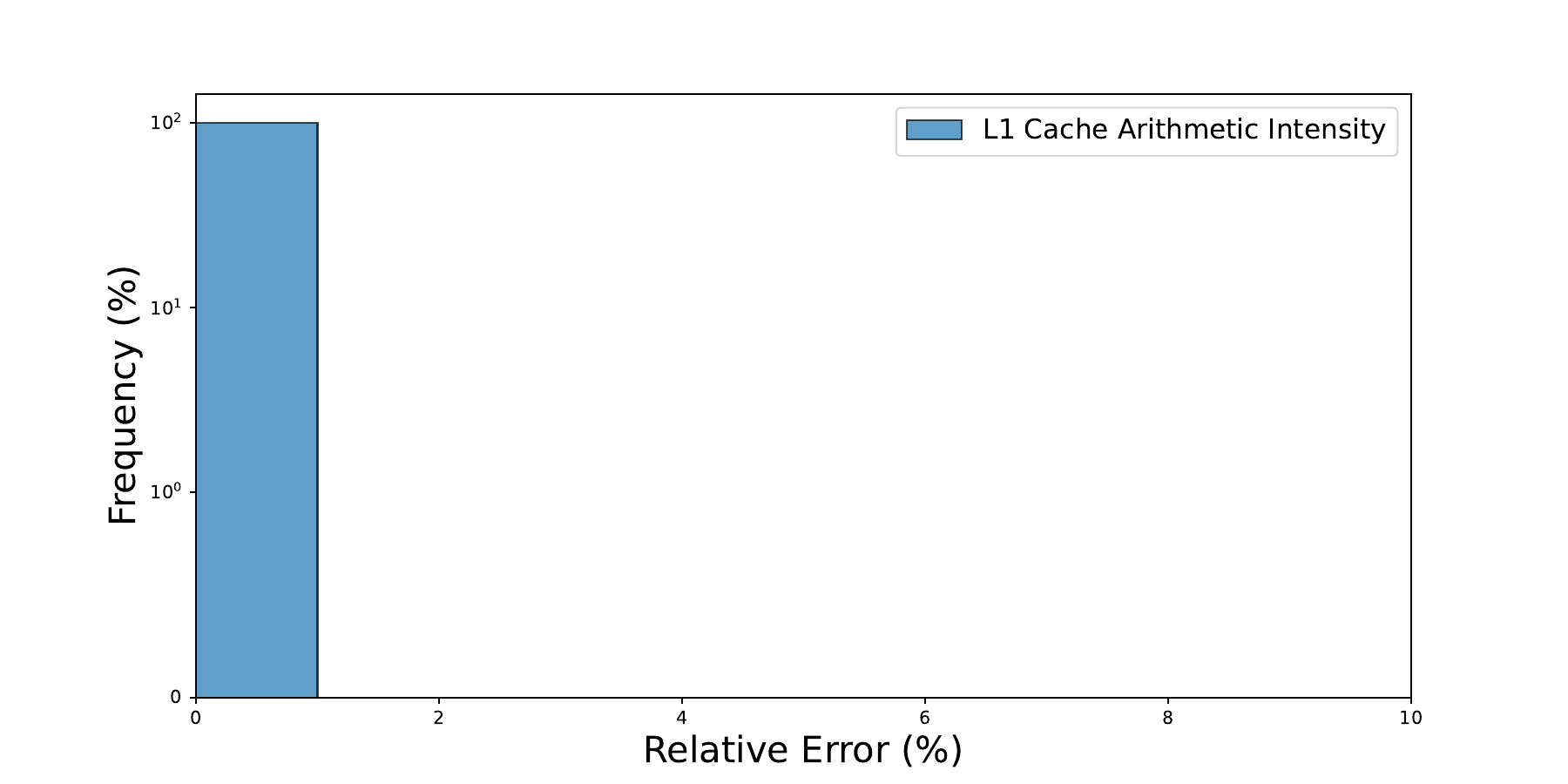}
        \caption{L1 Arithmetic Intensity - Relative error}
    \end{subfigure}
    \begin{subfigure}[t]{0.49\textwidth}
        \centering
        \includegraphics[width=1\linewidth]{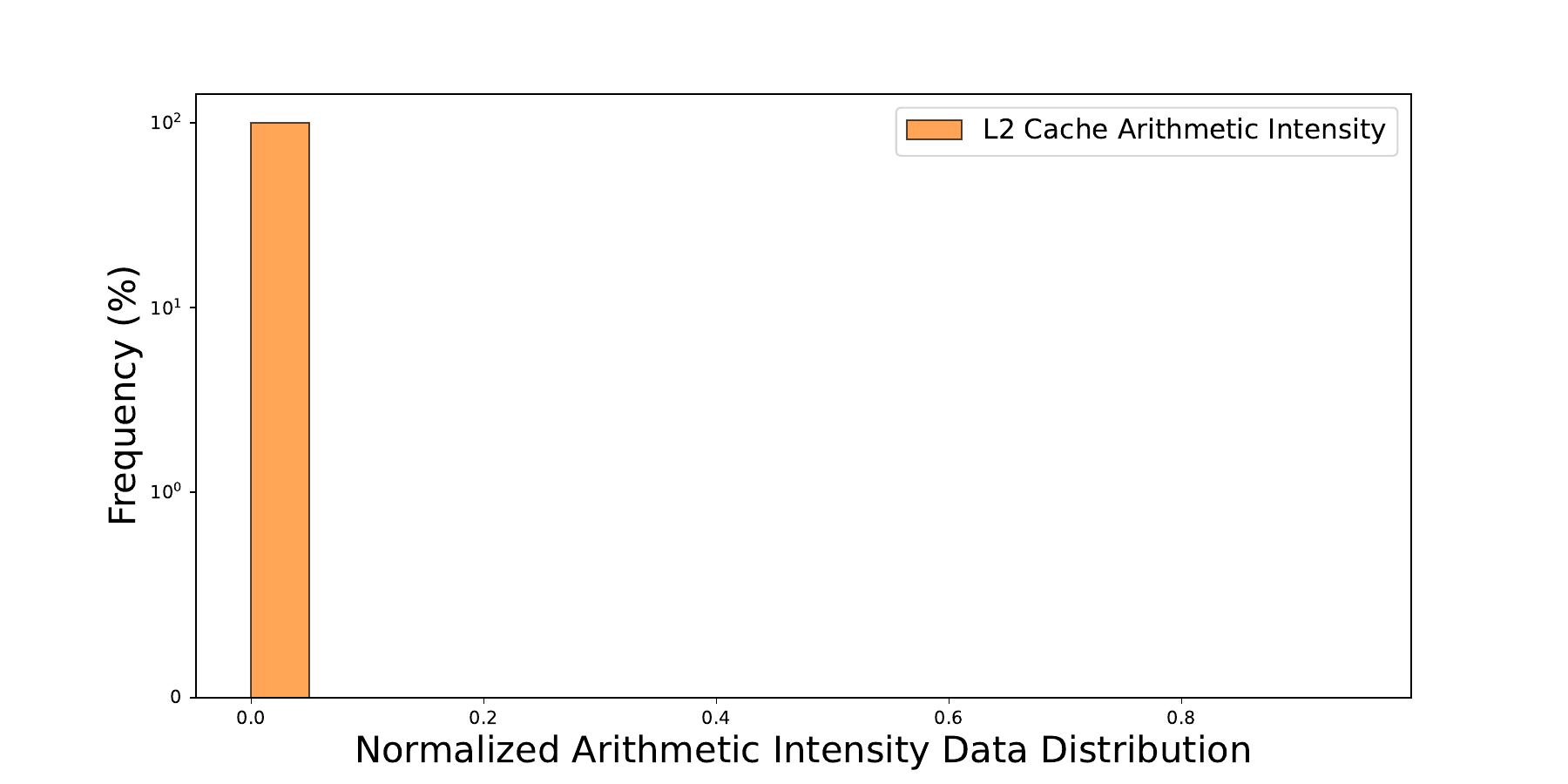}
        \caption{L2 Arithmetic Intensity - Input distribution}
    \end{subfigure}
    \begin{subfigure}[t]{0.49\textwidth}
        \centering
        \includegraphics[width=1\linewidth]{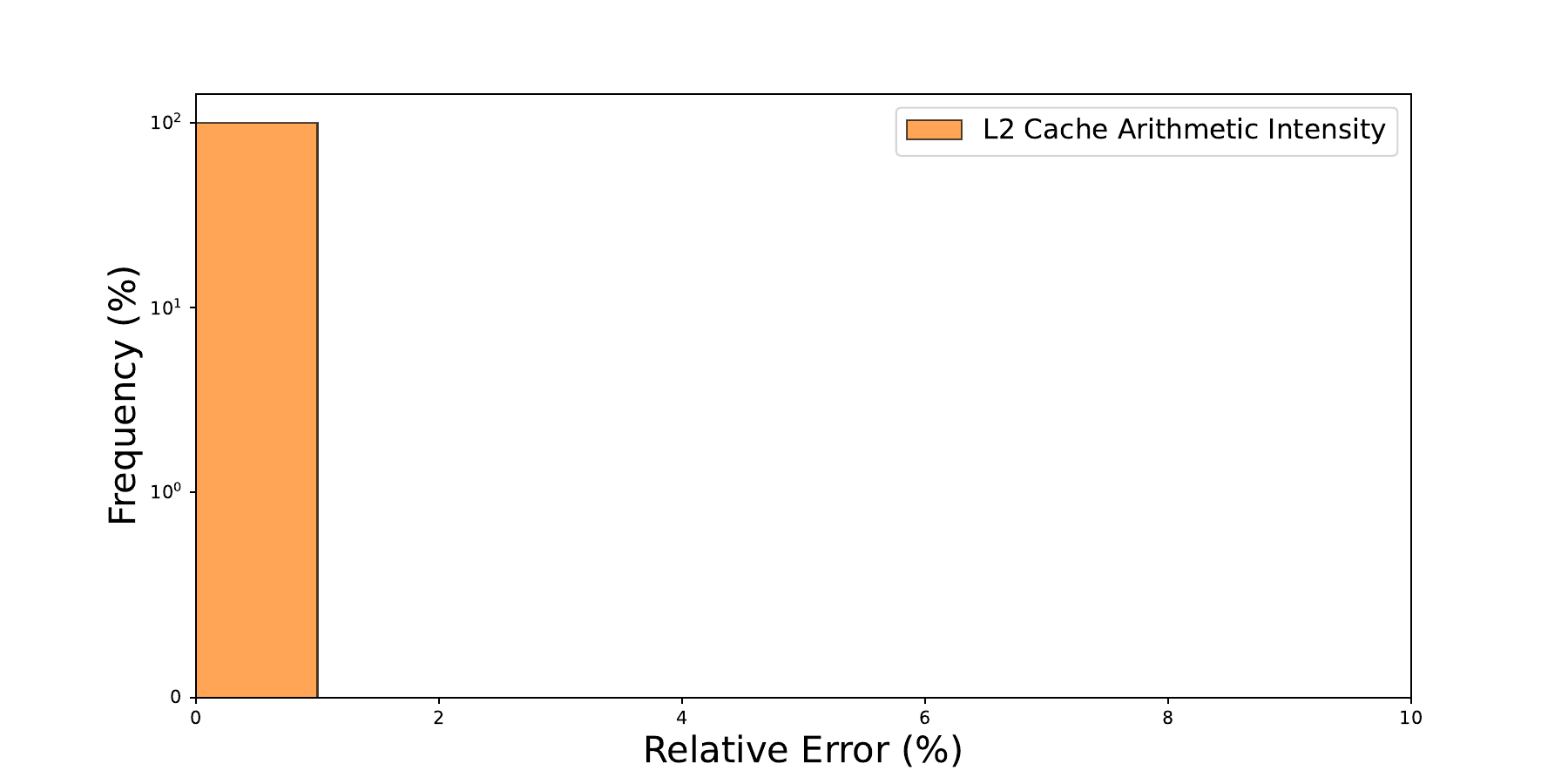}
        \caption{L2 Arithmetic Intensity - Relative error}
    \end{subfigure}
    \caption{Distributions and relative errors for L1 and L2 Arithmetic Intensity.}
    \label{fig:cache_ai_combined}
\end{figure*}

\begin{figure*}[p]
    \centering
    \begin{subfigure}[t]{0.49\textwidth}
        \centering
        \includegraphics[width=1\linewidth]{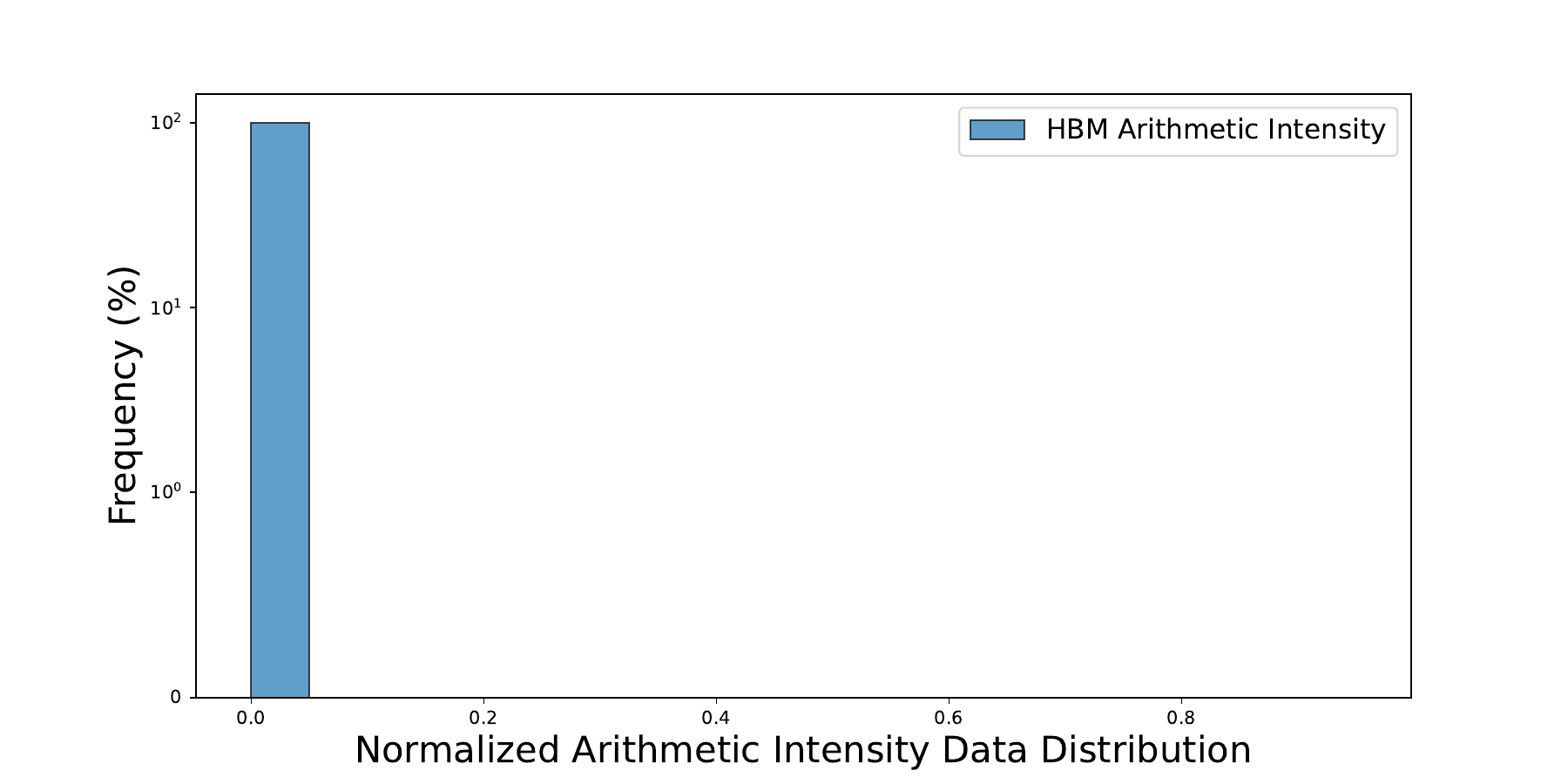}
        \caption{HBM Arithmetic Intensity - Input distribution}
    \end{subfigure}
    \begin{subfigure}[t]{0.49\textwidth}
        \centering
        \includegraphics[width=1\linewidth]{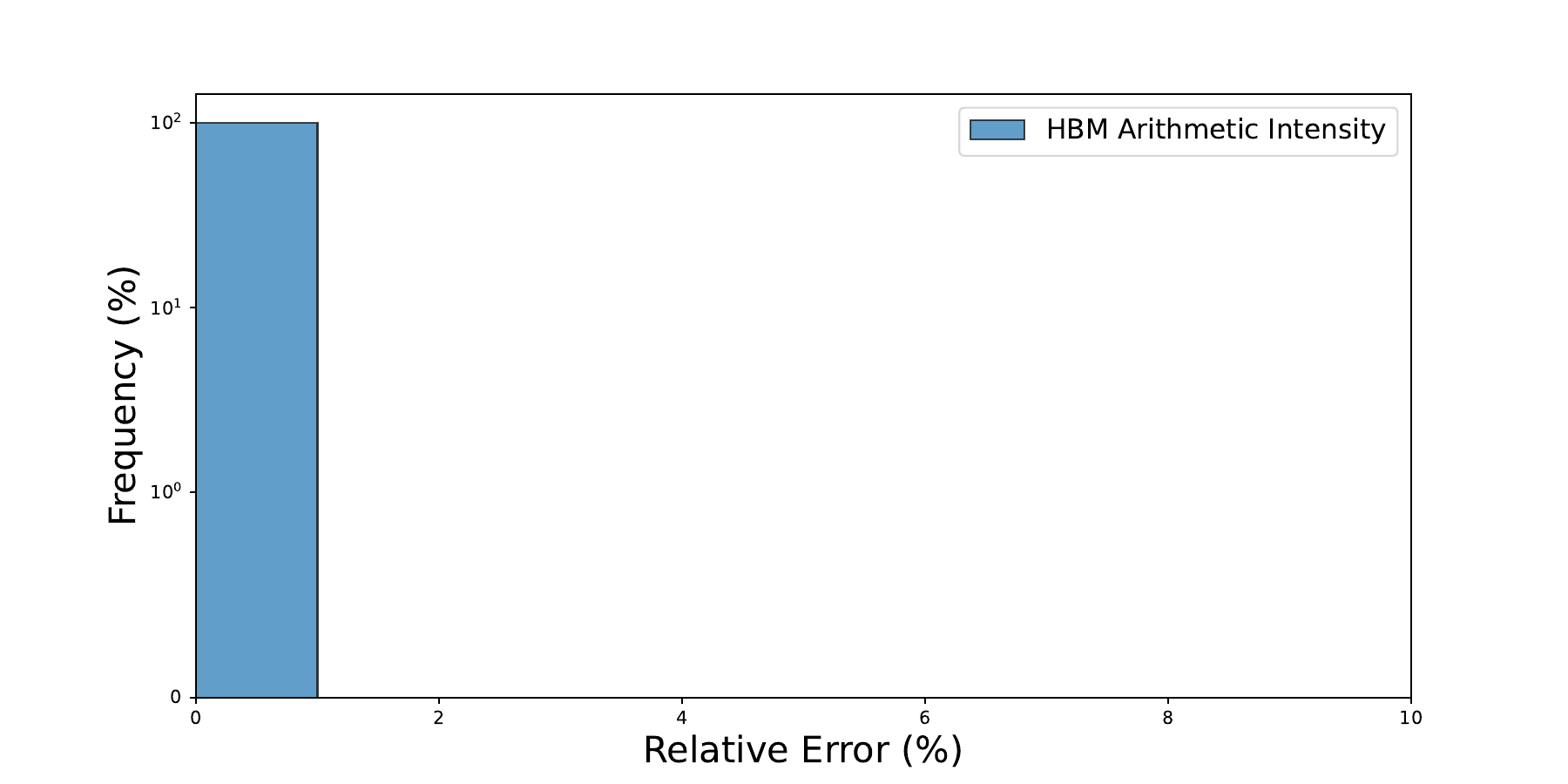}
        \caption{HBM Arithmetic Intensity - Relative error}
    \end{subfigure}
    \begin{subfigure}[t]{0.49\textwidth}
        \centering
        \includegraphics[width=1\linewidth]{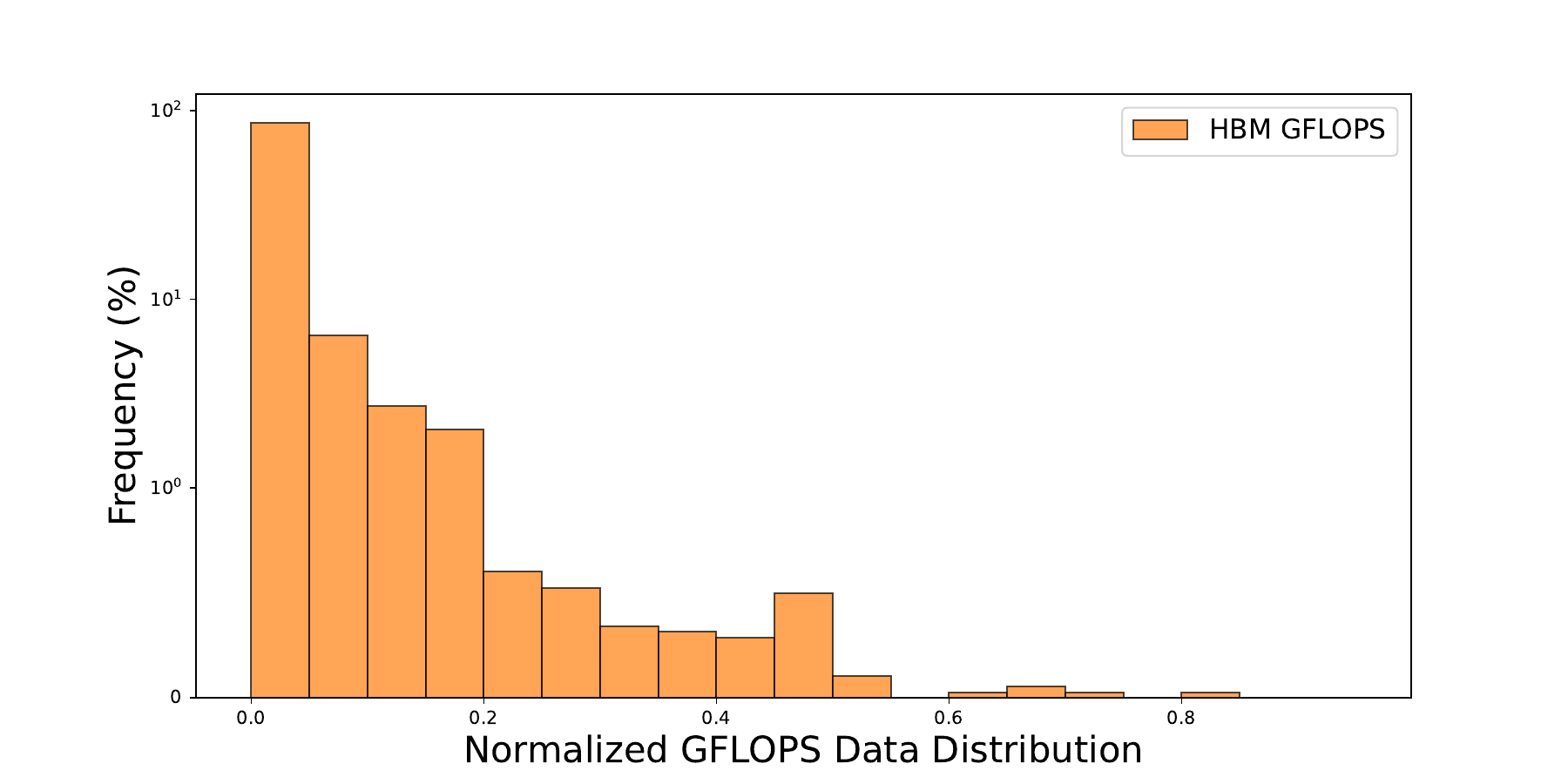}
        \caption{HBM GFLOP/s - Input distribution}
    \end{subfigure}
    \begin{subfigure}[t]{0.49\textwidth}
        \centering
        \includegraphics[width=1\linewidth]{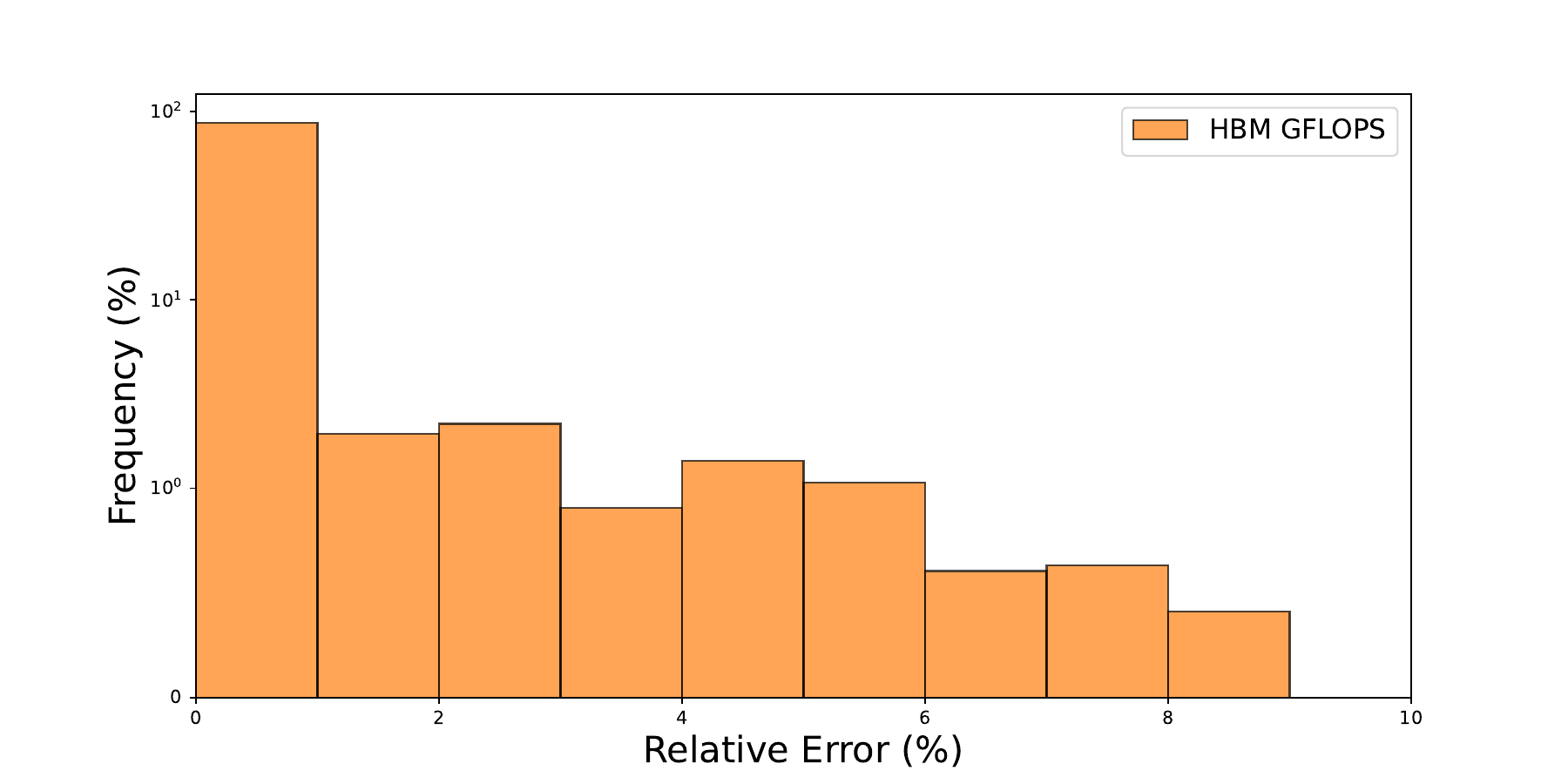}
        \caption{HBM GFLOP/s - Relative error}
    \end{subfigure}
    \caption{Distributions and relative errors for HBM Arithmetic Intensity and GFLOP/s.}
    \label{fig:hbm_combined}
\end{figure*}

\paragraph{Overall Performance.} 
As shown in Table~\ref{tab:cache_metrics}, \omniwise\ consistently achieves high predictive accuracy across all performance categories. Notably, over \textbf{90\% of all predictions fall within 10\% relative error}, even for metrics exhibiting significant input skew. For performance-critical counters such as arithmetic intensity and cache hit rate, \omniwise\ achieves \textbf{high accuracy}, with over 98.2\% and 87.3\% of predictions within 2\% error in Arithmetic Intensity and Cache Hit Rate predictions, respectively, as shown in Table~\ref{tab:cache_metrics} and Figures~\ref{fig:cache_hit_combined},~\ref{fig:cache_ai_combined}, and ~\ref{fig:hbm_combined}. These results validate the model's ability to learn non-trivial relationships between code semantics and hardware-level behavior, enabling practical prediction without any code execution.

\paragraph{Cache Hit Rates.} 
Figure~\ref{fig:cache_hit_combined} present data distributions for L1 and L2 cache hit rates. While the input data distributions exhibit moderate skew (e.g., L1 is concentrated near 0.55), the model produces highly accurate predictions, with \textbf{over 94.1\% within 10\% error} in L1 Cache Hit Rate and \textbf{over 89.9\% within just 2\%}. The model further captures diverse distribution of L2 Cache achieving 94.0\% within 10\% range and 87.3\% within 2\% range. These results are particularly meaningful because cache hit rates are notoriously sensitive to subtle variations in code and memory access patterns. The model's high fidelity across these metrics suggests it effectively captures latent memory access behavior from code syntax alone.

\begin{figure*}[p]
    \centering
    \begin{subfigure}[t]{0.49\textwidth}
        \centering
        \includegraphics[width=1\linewidth]{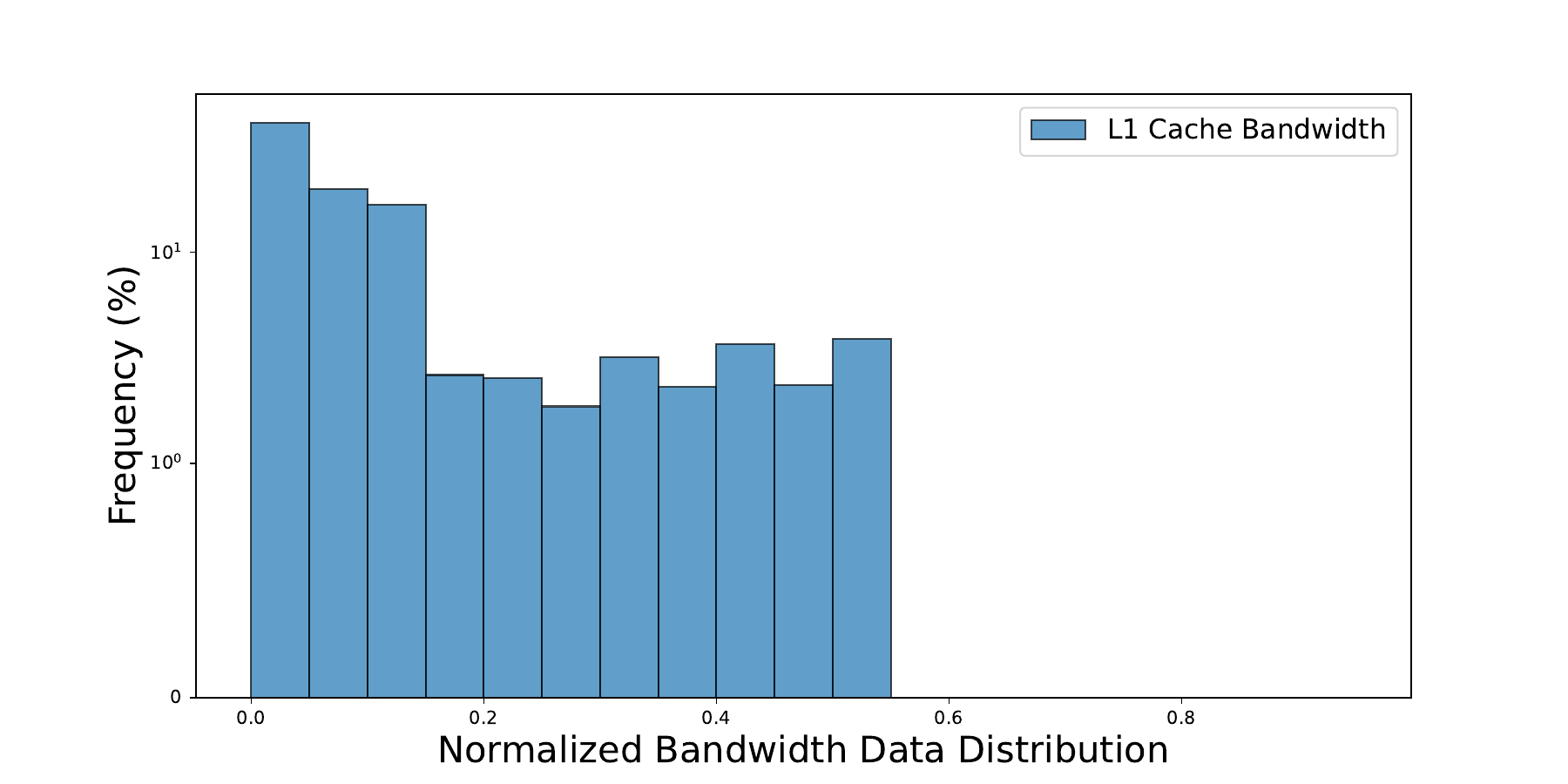}
        \caption{L1 Cache Bandwidth - Input distribution}
    \end{subfigure}
    \begin{subfigure}[t]{0.49\textwidth}
        \centering
        \includegraphics[width=1\linewidth]{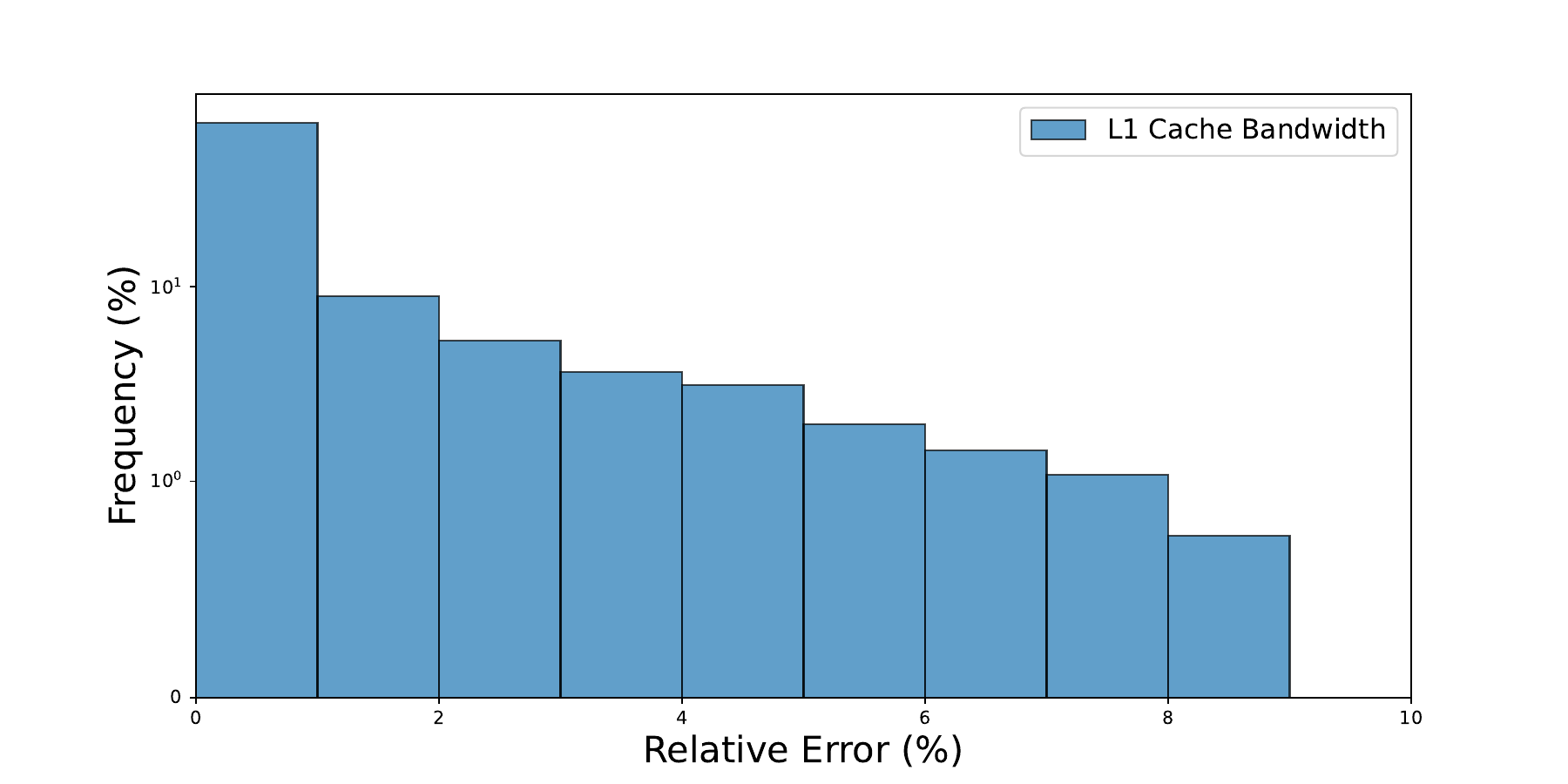}
        \caption{L1 Cache Bandwidth - Relative error}
    \end{subfigure}
    \begin{subfigure}[t]{0.49\textwidth}
        \centering
        \includegraphics[width=1\linewidth]{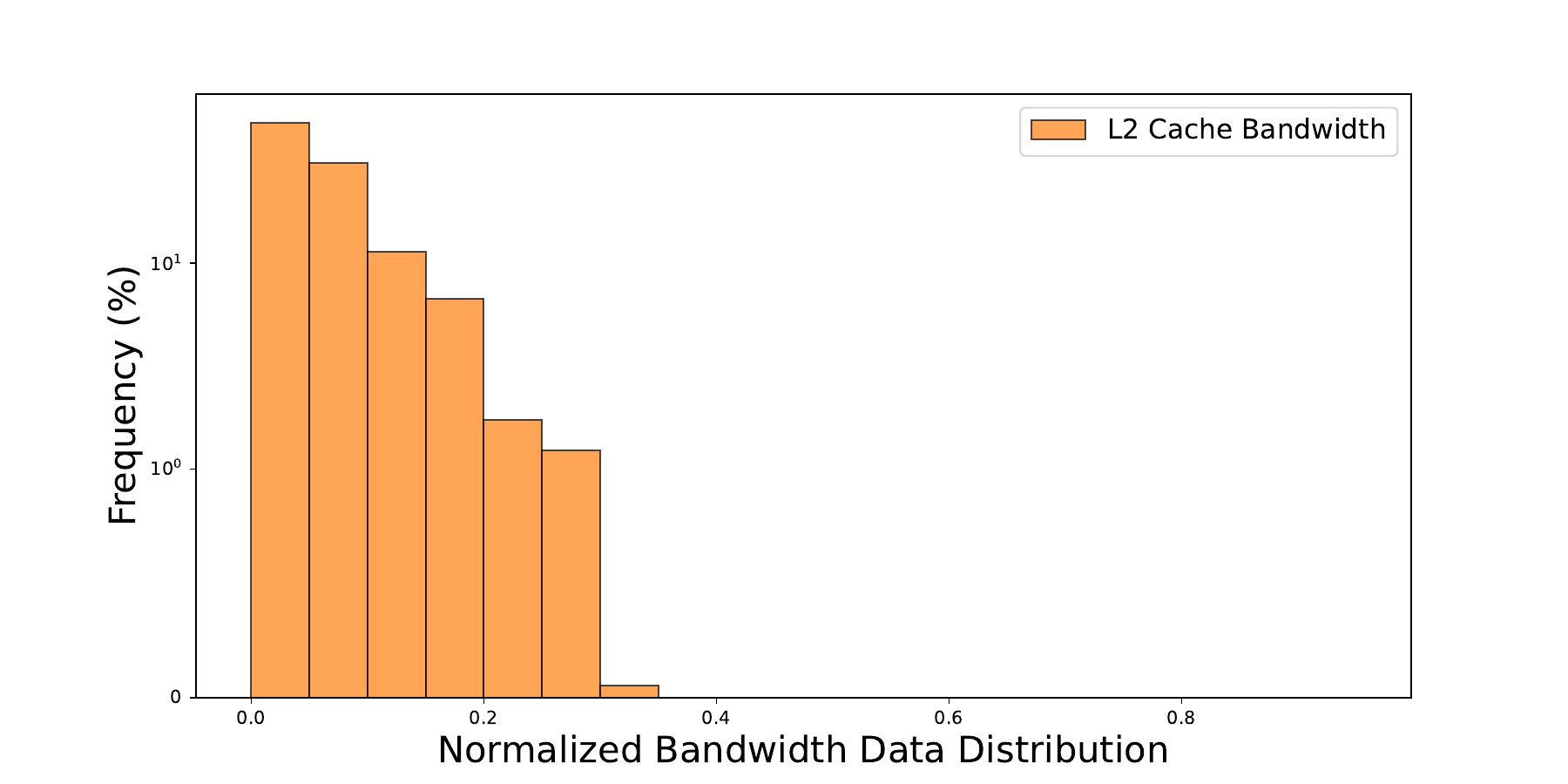}
        \caption{L2 Cache Bandwidth - Input distribution}
    \end{subfigure}
    \begin{subfigure}[t]{0.49\textwidth}
        \centering
        \includegraphics[width=1\linewidth]{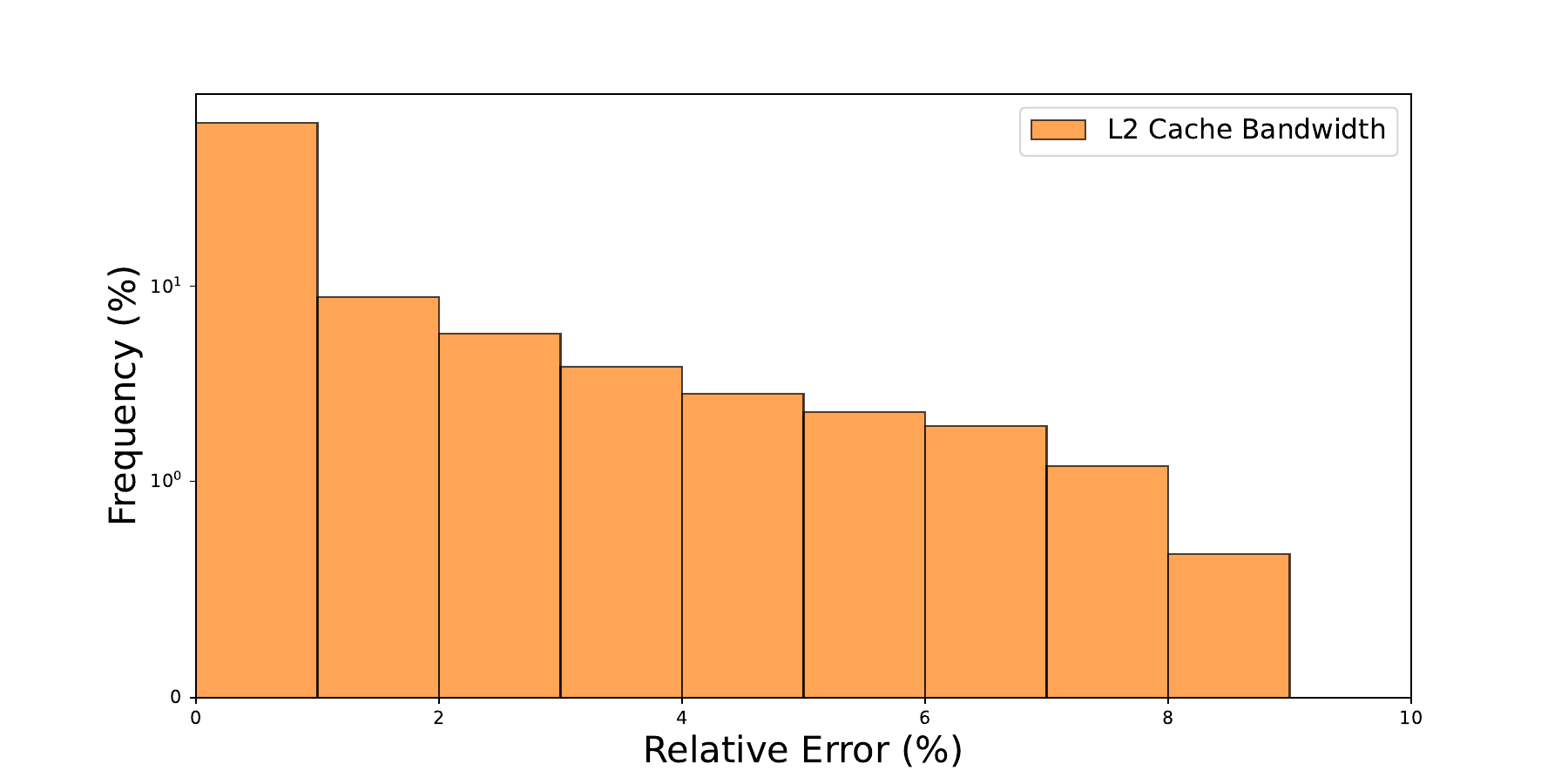}
        \caption{L2 Cache Bandwidth - Relative error}
    \end{subfigure}
    \caption{Distributions and relative errors for L1 and L2 Cache Bandwidth.}
    \label{fig:cache_bandwidth_combined}
\end{figure*}

\paragraph{Cache Bandwidth.} 
As illustrated in Figure~\ref{fig:cache_bandwidth_combined}, both L1 and L2 bandwidth distributions are heavily left-skewed, with most values near zero. Despite this skew, \omniwise\ maintains strong accuracy, achieving \textbf{90.8\% and 91.7\% within 10\% error} for L1 and L2 respectively. The ability to make such accurate predictions on sparse, non-Gaussian inputs highlights the model's robustness and generalization ability in memory throughput prediction.

\begin{figure*}[p]
    \centering
    \begin{subfigure}[t]{0.49\textwidth}
        \centering
        \includegraphics[width=1\linewidth]{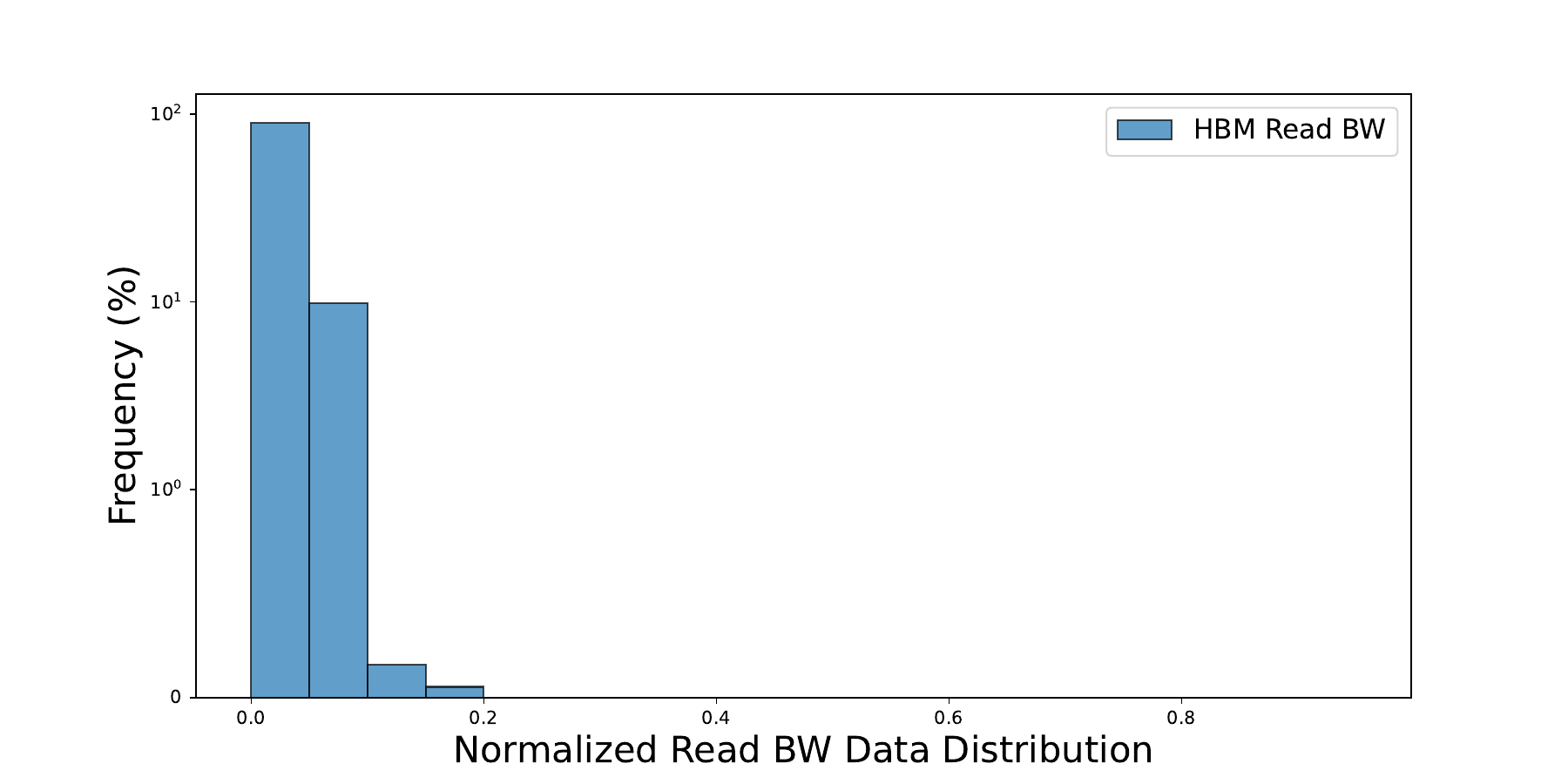}
        \caption{HBM Read BW - Input distribution}
    \end{subfigure}
    \begin{subfigure}[t]{0.49\textwidth}
        \centering
        \includegraphics[width=1\linewidth]{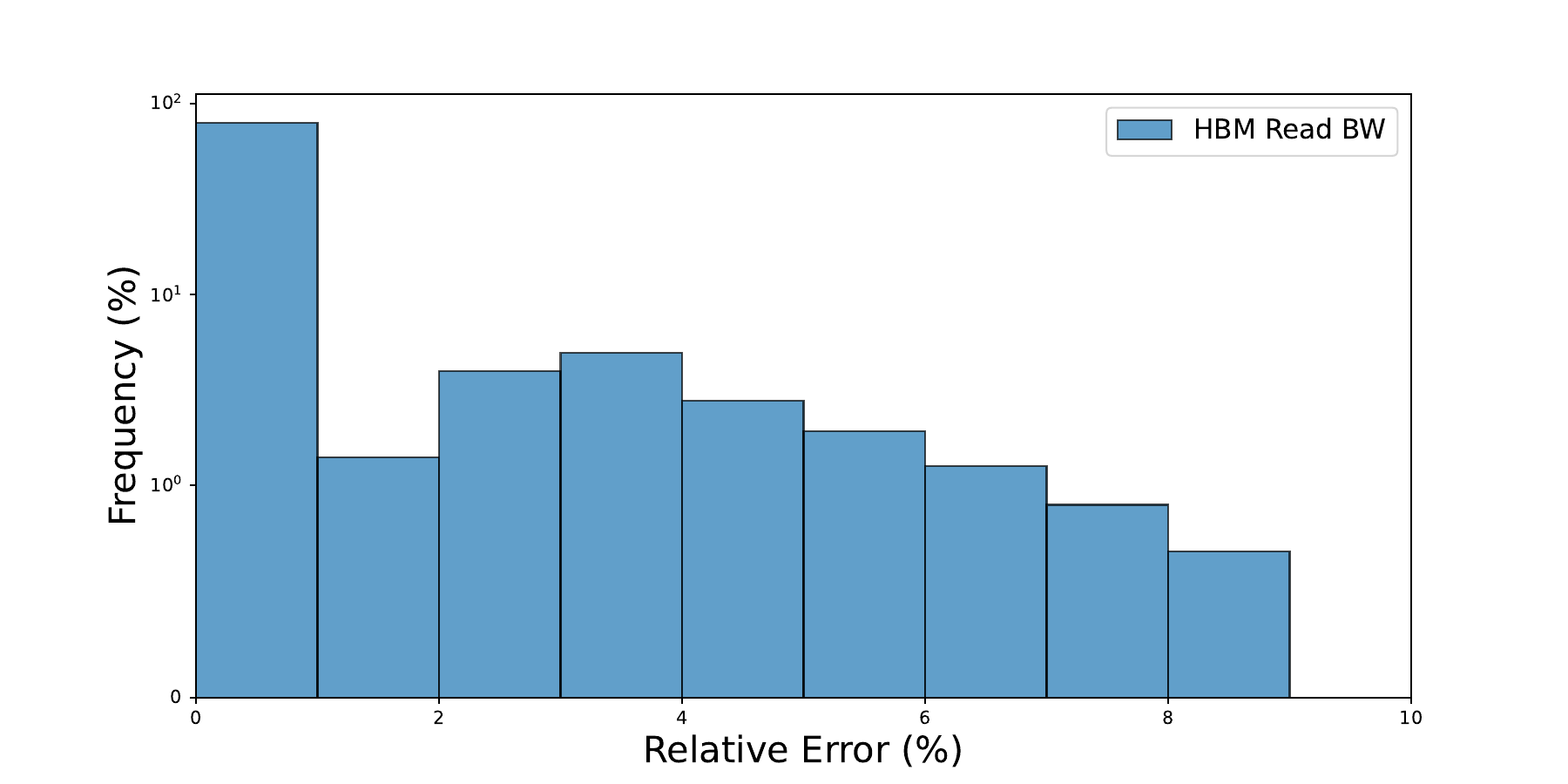}
        \caption{HBM Read BW - Relative error}
    \end{subfigure}
    \begin{subfigure}[t]{0.49\textwidth}
        \centering
        \includegraphics[width=1\linewidth]{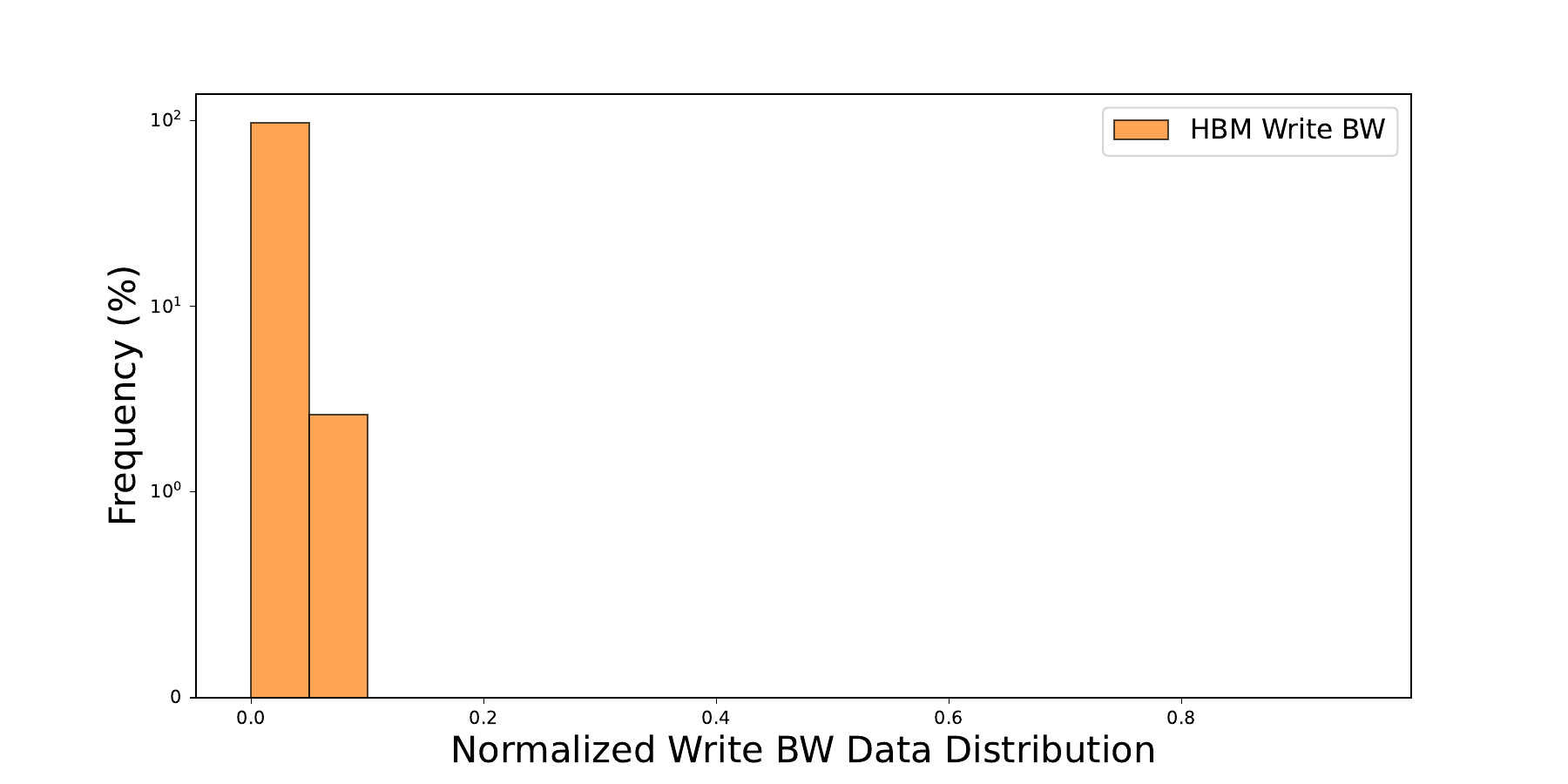}
        \caption{HBM Write BW - Input distribution}
    \end{subfigure}
    \begin{subfigure}[t]{0.49\textwidth}
        \centering
        \includegraphics[width=1\linewidth]{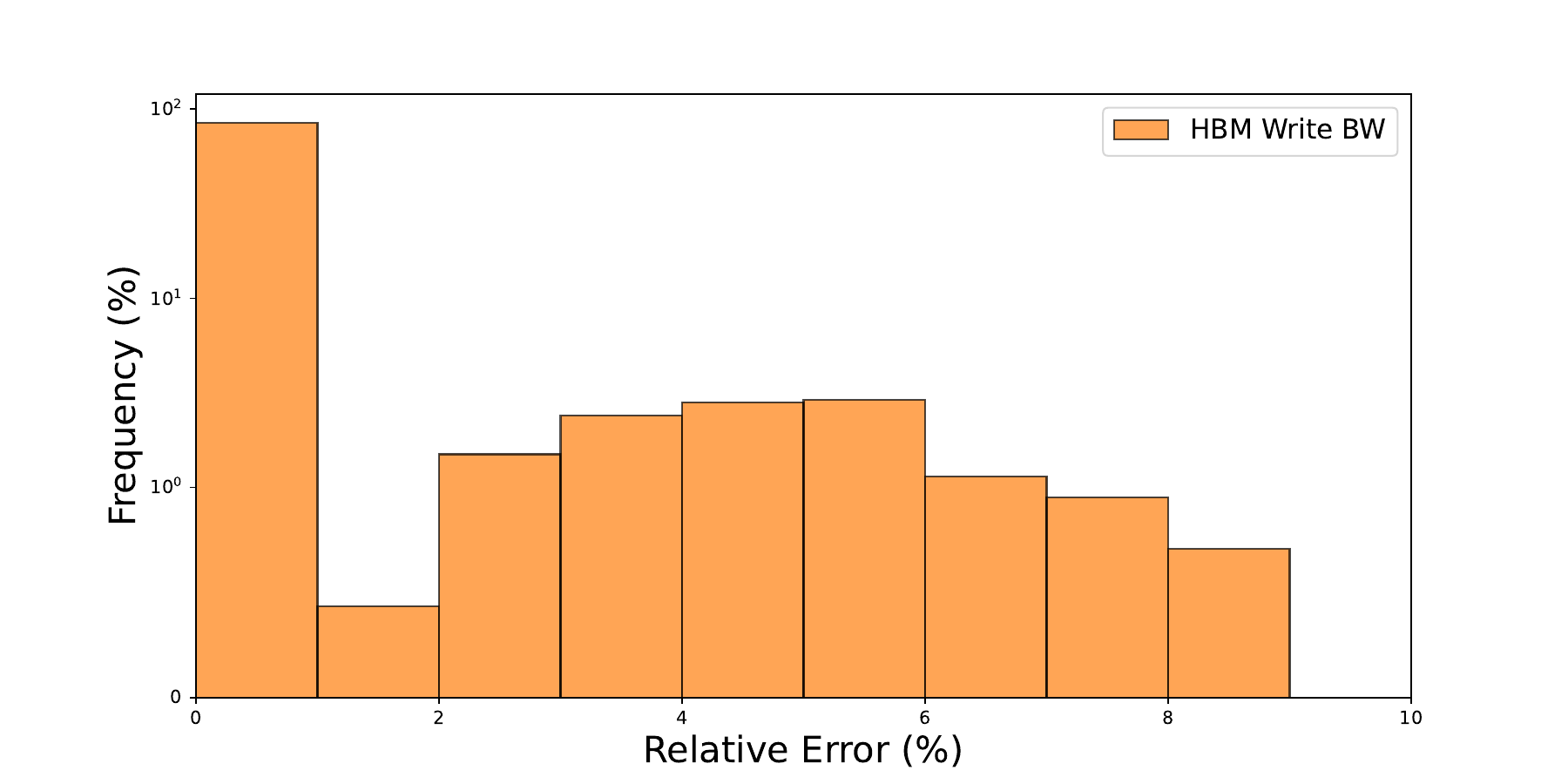}
        \caption{HBM Write BW - Relative error}
    \end{subfigure}
    \caption{Distributions and relative errors for HBM Read and Write Bandwidth.}
    \label{fig:fabric_bw_combined}
\end{figure*}

\paragraph{HBM Read \& Write Bandwidth.} 
Figure~\ref{fig:fabric_bw_combined} further pushes the boundary of prediction under extreme data imbalance. Both read and write bandwidths are concentrated near zero in the training set, with over 95\% of values below 0.05. Nonetheless, the model achieves \textbf{92.9\% (read) and 93.8\% (write)} of predictions within 10\% error. These results underscore the effectiveness of normalization and token formatting strategies, and demonstrate that even sparsely active metrics can be accurately predicted without compromising numerical resolution.

\paragraph{Arithmetic Intensity.} 
Across all three memory levels—L1, L2, and HBM—\omniwise\ demonstrates \textbf{virtually perfect generalization} for arithmetic intensity metrics. As shown in Figures \ref{fig:cache_ai_combined} and \ref{fig:hbm_combined}, more than 99.8\% of predictions fall within 2\% error across the board. Given that arithmetic intensity correlates with compute-bound kernel behavior, this level of precision supports the use of \omniwise\ not just for profiling, but also for static performance diagnosis, design-space exploration, and compute efficiency tuning.

\paragraph{HBM GFLOPs.} 
While more volatile than intensity metrics, predictions on HBM GFLOPs also perform well. Despite strong skew in the input distribution (Figure~\ref{fig:hbm_combined}), the model achieves \textbf{91.7\% of predictions within 10\% error}. This highlights the pipeline's ability to generalize beyond cache behavior and into raw throughput estimation, which is often used as a primary indicator of kernel-level performance.

\paragraph{Discussion.} 
These results collectively demonstrate that \omniwise\ is not only effective, but also \textbf{practical and generalizable}. The model's performance holds across highly imbalanced, hardware-sensitive metrics—an area where conventional static analyzers, heuristics, or analytic roofline models tend to break down. By predicting fine-grained hardware counters from code alone, \omniwise\ enables developers to reason about performance bottlenecks, memory saturation, and compute intensity in real-time, \textbf{without code execution or instrumentation overhead}. This suggests a compelling direction for LLM-powered performance diagnostics at scale.


%% file: tex/conclusion.tex
\section{Conclusion and Future Work}
\label{sec:conclusion}

We presented \omniwise{}, an end-to-end pipeline designed to aid GPU programmers to write efficient GPU kernels. Our pipeline starts from data collection and curation all the way to model serving in a developer-friendly Visual Studio Code extension. \omniwise{} achieves more than 90\% accuracy predicting memory and compute metrics, allowing developers to gain an idea on how their code is performing without having to wait for long development cycles. It is essential to understand that \omniwise is designed to predict performance counters for GPU kernels using particular large problem sizes. When users are working with large problem sizes, the predicted counters shall apply to the source code; however, when the problem sizes differ, users need to have some idea of which performance counters are invariant to the problem size (for instance, counters that are directly related to algorithmic complexity).

We believe that the methodology developed in \omniwise{} can be further extended and applied to various problems  and other programming languages beyond HIP\@. Our future plans, include targeting other high and intermediate languages, broadening our predicted set of counters as well as further handling various problem sizes, launch bounds and data types.

Furthermore, in subsequent versions of \omniwise{}, we intend to incorporate additional QA datasets to further refine the model's capacity to understand and adapt to diverse code structures. This will enhance our ability to handle more complex real-world scenarios and codebases. Alongside this dataset expansion, we also plan to employ reinforcement learning techniques to iteratively improve \omniwise{}'s predictions. Through an RL-driven feedback loop, the model can learn from its own performance outcomes and optimize its parameters accordingly, thereby continually boosting accuracy and robustness.

Taken together, these future directions are designed to make \omniwise{} more versatile, scalable, and adaptive, ensuring that our platform remains at the forefront of performance analysis and optimization in ever-evolving computing environments.

%% file: tex/acknowledgments.tex
\begin{acks}
This work was supported in part by Advanced Micro Devices, Inc. under the AMD AI \& HPC Cluster Program. The authors would like to thank Karl Schulz, Jianbing Chen, Michael Schulte and Ralph Wittig for their continuous feedback, support and suggestions to improve \omniwise{}. AMD, the AMD Arrow logo, AMD CDNA, AMD Instinct, AMD ROCm,
AMD Infinity Cache, AMD Infinity Fabric, and combinations
thereof are trademarks of Advanced Micro Devices, Inc\@. Other
product names used in this publication are for identification
purposes only and may be trademarks of their respective
companies.

\end{acks}

%% file: omniwise.bbl

\begin{thebibliography}{25}


\ifx \showCODEN    \undefined \def \showCODEN     #1{\unskip}     \fi
\ifx \showISBNx    \undefined \def \showISBNx     #1{\unskip}     \fi
\ifx \showISBNxiii \undefined \def \showISBNxiii  #1{\unskip}     \fi
\ifx \showISSN     \undefined \def \showISSN      #1{\unskip}     \fi
\ifx \showLCCN     \undefined \def \showLCCN      #1{\unskip}     \fi
\ifx \shownote     \undefined \def \shownote      #1{#1}          \fi
\ifx \showarticletitle \undefined \def \showarticletitle #1{#1}   \fi
\ifx \showURL      \undefined \def \showURL       {\relax}        \fi
\providecommand\bibfield[2]{#2}
\providecommand\bibinfo[2]{#2}
\providecommand\natexlab[1]{#1}
\providecommand\showeprint[2][]{arXiv:#2}

\bibitem[Achiam et~al\mbox{.}(2023)]%
        {Achiam:2023:GTR}
\bibfield{author}{\bibinfo{person}{Josh Achiam}, \bibinfo{person}{Steven
  Adler}, \bibinfo{person}{Sandhini Agarwal}, \bibinfo{person}{Lama Ahmad},
  \bibinfo{person}{Ilge Akkaya}, \bibinfo{person}{Florencia~Leoni Aleman},
  \bibinfo{person}{Diogo Almeida}, \bibinfo{person}{Janko Altenschmidt},
  \bibinfo{person}{Sam Altman}, \bibinfo{person}{Shyamal Anadkat},
  {et~al\mbox{.}}} \bibinfo{year}{2023}\natexlab{}.
\newblock \showarticletitle{{GPT}-4 Technical Report}.
\newblock \bibinfo{journal}{\emph{arXiv preprint arXiv:2303.08774}}
  (\bibinfo{year}{2023}).
\newblock
\href{https://doi.org/10.48550/arXiv.2303.08774}{doi:\nolinkurl{10.48550/arXiv.2303.08774}}


\bibitem[{AI}(2025)]%
        {HF:2024:L32}
\bibfield{author}{\bibinfo{person}{Meta {AI}}.}
  \bibinfo{year}{2025}\natexlab{}.
\newblock \bibinfo{title}{{LLaMA} 3.2-3B Instruct}.
\newblock
  \bibinfo{howpublished}{\url{https://huggingface.co/meta-llama/Llama-3.2-3B-Instruct}}.
\newblock
\newblock
\shownote{[Online; accessed 17-January-2025]}.


\bibitem[{AMD}(2025)]%
        {AMD:2025:HTE}
\bibfield{author}{\bibinfo{person}{{AMD}}.} \bibinfo{year}{2025}\natexlab{}.
\newblock \bibinfo{title}{{HPC} Training Examples}.
\newblock
  \bibinfo{howpublished}{\url{https://github.com/amd/HPCTrainingExamples}}.
\newblock
\newblock
\shownote{[Online; accessed 17-January-2025]}.


\bibitem[Awad et~al\mbox{.}(2023)]%
        {Awad:2023:AAI}
\bibfield{author}{\bibinfo{person}{Muhammad~A. Awad}, \bibinfo{person}{Saman
  Ashkiani}, \bibinfo{person}{Serban~D. Porumbescu},
  \bibinfo{person}{Mart{\'{i}}n Farach-Colton}, {and} \bibinfo{person}{John~D.
  Owens}.} \bibinfo{year}{2023}\natexlab{}.
\newblock \showarticletitle{Analyzing and Implementing {GPU} Hash Tables}. In
  \bibinfo{booktitle}{\emph{SIAM Symposium on Algorithmic Principles of
  Computer Systems}} \emph{(\bibinfo{series}{APOCS23})}.
  \bibinfo{pages}{33--50}.
\newblock
\href{https://doi.org/10.1137/1.9781611977578.ch3}{doi:\nolinkurl{10.1137/1.9781611977578.ch3}}


\bibitem[Corporation(2024)]%
        {NVIDIA:2024:NNC}
\bibfield{author}{\bibinfo{person}{NVIDIA Corporation}.}
  \bibinfo{year}{2024}\natexlab{}.
\newblock \bibinfo{booktitle}{\emph{NVIDIA Nsight Compute}}.
\newblock
\newblock
\shownote{Accessed: 2024-11-05}.


\bibitem[Cummins et~al\mbox{.}(2024)]%
        {Cummins:2024:MLM}
\bibfield{author}{\bibinfo{person}{Chris Cummins}, \bibinfo{person}{Volker
  Seeker}, \bibinfo{person}{Dejan Grubisic}, \bibinfo{person}{Baptiste
  Roziere}, \bibinfo{person}{Jonas Gehring}, \bibinfo{person}{Gabriel
  Synnaeve}, {and} \bibinfo{person}{Hugh Leather}.}
  \bibinfo{year}{2024}\natexlab{}.
\newblock \bibinfo{title}{Meta Large Language Model Compiler: Foundation Models
  of Compiler Optimization}.
\newblock
\showeprint[arxiv]{2407.02524}~[cs.PL]
\href{https://doi.org/10.48550/arXiv.2407.02524}{doi:\nolinkurl{10.48550/arXiv.2407.02524}}


\bibitem[Dubey et~al\mbox{.}(2024)]%
        {Dubey:2024:L3H}
\bibfield{author}{\bibinfo{person}{Abhimanyu Dubey}, \bibinfo{person}{Abhinav
  Jauhri}, \bibinfo{person}{Abhinav Pandey}, \bibinfo{person}{Abhishek Kadian},
  \bibinfo{person}{Ahmad Al-Dahle}, \bibinfo{person}{Aiesha Letman},
  \bibinfo{person}{Akhil Mathur}, \bibinfo{person}{Alan Schelten},
  \bibinfo{person}{Amy Yang}, \bibinfo{person}{Angela Fan}, {et~al\mbox{.}}}
  \bibinfo{year}{2024}\natexlab{}.
\newblock \showarticletitle{The {LLaMA} 3 Herd of Models}.
\newblock \bibinfo{journal}{\emph{arXiv preprint arXiv:2407.21783}}
  (\bibinfo{year}{2024}).
\newblock
\href{https://doi.org/10.48550/arXiv.2407.21783}{doi:\nolinkurl{10.48550/arXiv.2407.21783}}


\bibitem[Feng et~al\mbox{.}(2020)]%
        {Feng:2020:CAP}
\bibfield{author}{\bibinfo{person}{Zhangyin Feng}, \bibinfo{person}{Daya Guo},
  \bibinfo{person}{Duyu Tang}, \bibinfo{person}{Nan Duan},
  \bibinfo{person}{Xiaocheng Feng}, \bibinfo{person}{Ming Gong},
  \bibinfo{person}{Linjun Shou}, \bibinfo{person}{Bing Qin},
  \bibinfo{person}{Ting Liu}, \bibinfo{person}{Daxin Jiang}, {et~al\mbox{.}}}
  \bibinfo{year}{2020}\natexlab{}.
\newblock \showarticletitle{{CodeBERT}: A Pre-Trained Model for Programming and
  Natural Languages}.
\newblock \bibinfo{journal}{\emph{arXiv preprint arXiv:2002.08155}}
  (\bibinfo{year}{2020}).
\newblock
\href{https://doi.org/10.48550/arXiv.2002.08155}{doi:\nolinkurl{10.48550/arXiv.2002.08155}}


\bibitem[Gupta et~al\mbox{.}(2012)]%
        {Gupta:2012:ASO}
\bibfield{author}{\bibinfo{person}{Kshitij Gupta}, \bibinfo{person}{Jeff
  Stuart}, {and} \bibinfo{person}{John~D. Owens}.}
  \bibinfo{year}{2012}\natexlab{}.
\newblock \showarticletitle{A Study of Persistent Threads Style {GPU}
  Programming for {GPGPU} Workloads}. In \bibinfo{booktitle}{\emph{Proceedings
  of Innovative Parallel Computing}} \emph{(\bibinfo{series}{InPar '12})}.
\newblock
\href{https://doi.org/10.1109/InPar.2012.6339596}{doi:\nolinkurl{10.1109/InPar.2012.6339596}}


\bibitem[Hu et~al\mbox{.}(2021)]%
        {Hu:2021:LLR}
\bibfield{author}{\bibinfo{person}{Edward~J. Hu}, \bibinfo{person}{Yelong
  Shen}, \bibinfo{person}{Phillip Wallis}, \bibinfo{person}{Zeyuan Allen-Zhu},
  \bibinfo{person}{Yuanzhi Li}, \bibinfo{person}{Shean Wang},
  \bibinfo{person}{Lu Wang}, {and} \bibinfo{person}{Weizhu Chen}.}
  \bibinfo{year}{2021}\natexlab{}.
\newblock \showarticletitle{{LoRA}: Low-Rank Adaptation of Large Language
  Models}.
\newblock \bibinfo{journal}{\emph{arXiv preprint arXiv:2106.09685}}
  (\bibinfo{year}{2021}).
\newblock
\href{https://doi.org/10.48550/arXiv.2106.09685}{doi:\nolinkurl{10.48550/arXiv.2106.09685}}


\bibitem[Kingma et~al\mbox{.}(2020)]%
        {adam}
\bibfield{author}{\bibinfo{person}{Diederik~P Kingma}, \bibinfo{person}{J~Adam
  Ba}, {and} \bibinfo{person}{J Adam}.} \bibinfo{year}{2020}\natexlab{}.
\newblock \showarticletitle{A method for stochastic optimization. arXiv 2014}.
\newblock \bibinfo{journal}{\emph{arXiv preprint arXiv:1412.6980}}
  \bibinfo{volume}{106} (\bibinfo{year}{2020}), \bibinfo{pages}{6}.
\newblock


\bibitem[Lozhkov et~al\mbox{.}(2024)]%
        {Lozhkov:2024:SSC}
\bibfield{author}{\bibinfo{person}{Anton Lozhkov}, \bibinfo{person}{Raymond
  Li}, \bibinfo{person}{Loubna~Ben Allal}, \bibinfo{person}{Federico Cassano},
  \bibinfo{person}{Joel Lamy-Poirier}, \bibinfo{person}{Nouamane Tazi},
  \bibinfo{person}{Ao Tang}, \bibinfo{person}{Dmytro Pykhtar},
  \bibinfo{person}{Jiawei Liu}, \bibinfo{person}{Yuxiang Wei}, {et~al\mbox{.}}}
  \bibinfo{year}{2024}\natexlab{}.
\newblock \showarticletitle{{StarCoder} 2 and The Stack v2: The Next
  Generation}.
\newblock \bibinfo{journal}{\emph{arXiv preprint arXiv:2402.19173}}
  (\bibinfo{year}{2024}).
\newblock
\href{https://doi.org/10.48550/arXiv.2402.19173}{doi:\nolinkurl{10.48550/arXiv.2402.19173}}


\bibitem[Lu et~al\mbox{.}(2024)]%
        {Lu:2024:RRC}
\bibfield{author}{\bibinfo{person}{Xiaomin Lu}, \bibinfo{person}{Cole Ramos},
  \bibinfo{person}{Fei Zheng}, \bibinfo{person}{Karl~W. Schulz},
  \bibinfo{person}{Jose Santos}, \bibinfo{person}{Keith Lowery},
  \bibinfo{person}{Nicholas Curtis}, {and} \bibinfo{person}{Cristian~Di
  Pietrantonio}.} \bibinfo{year}{2024}\natexlab{}.
\newblock \bibinfo{booktitle}{\emph{ROCm/rocprofiler-compute: v3.0.0 (01
  November 2024)}}.
\newblock
\href{https://doi.org/10.5281/zenodo.7314631}{doi:\nolinkurl{10.5281/zenodo.7314631}}


\bibitem[{Meta AI}(2025)]%
        {Meta:2024:L32}
\bibfield{author}{\bibinfo{person}{{Meta AI}}.}
  \bibinfo{year}{2025}\natexlab{}.
\newblock \bibinfo{title}{{LLaMA} 3: 2024 Vision for {Edge} and {Mobile}
  Devices}.
\newblock
  \bibinfo{howpublished}{\url{https://ai.meta.com/blog/llama-3-2-connect-2024-vision-edge-mobile-devices/}}.
\newblock
\newblock
\shownote{[Online; accessed 17-January-2025]}.


\bibitem[OpenAI(2025)]%
        {OpenAI:2024:HGO}
\bibfield{author}{\bibinfo{person}{OpenAI}.} \bibinfo{year}{2025}\natexlab{}.
\newblock \bibinfo{title}{Hello {GPT}-4o}.
\newblock \bibinfo{howpublished}{\url{https://openai.com/index/hello-gpt-4o/}}.
\newblock
\newblock
\shownote{[Online; accessed 17-January-2025]}.


\bibitem[Rajbhandari et~al\mbox{.}(2020)]%
        {Rajbhandari:2020:ZMO}
\bibfield{author}{\bibinfo{person}{Samyam Rajbhandari}, \bibinfo{person}{Jeff
  Rasley}, \bibinfo{person}{Olatunji Ruwase}, {and} \bibinfo{person}{Yuxiong
  He}.} \bibinfo{year}{2020}\natexlab{}.
\newblock \showarticletitle{{ZeRO}: Memory Optimizations Toward Training
  Trillion Parameter Models}. In \bibinfo{booktitle}{\emph{SC20: International
  Conference for High Performance Computing, Networking, Storage and
  Analysis}}. \bibinfo{pages}{1--16}.
\newblock
\href{https://doi.org/10.1109/SC41405.2020.00024}{doi:\nolinkurl{10.1109/SC41405.2020.00024}}


\bibitem[Rasley et~al\mbox{.}(2020)]%
        {rasley2020deepspeed}
\bibfield{author}{\bibinfo{person}{Jeff Rasley}, \bibinfo{person}{Samyam
  Rajbhandari}, \bibinfo{person}{Olatunji Ruwase}, {and}
  \bibinfo{person}{Yuxiong He}.} \bibinfo{year}{2020}\natexlab{}.
\newblock \showarticletitle{Deepspeed: System optimizations enable training
  deep learning models with over 100 billion parameters}. In
  \bibinfo{booktitle}{\emph{Proceedings of the 26th ACM SIGKDD international
  conference on knowledge discovery \& data mining}}.
  \bibinfo{pages}{3505--3506}.
\newblock


\bibitem[Reddi et~al\mbox{.}(2020)]%
        {reddi2020mlperf}
\bibfield{author}{\bibinfo{person}{Vijay~Janapa Reddi},
  \bibinfo{person}{Christine Cheng}, \bibinfo{person}{David Kanter},
  \bibinfo{person}{Peter Mattson}, \bibinfo{person}{Guenther Schmuelling},
  \bibinfo{person}{Carole-Jean Wu}, \bibinfo{person}{Brian Anderson},
  \bibinfo{person}{Maximilien Breughe}, \bibinfo{person}{Mark Charlebois},
  \bibinfo{person}{William Chou}, {et~al\mbox{.}}}
  \bibinfo{year}{2020}\natexlab{}.
\newblock \showarticletitle{Mlperf inference benchmark}. In
  \bibinfo{booktitle}{\emph{2020 ACM/IEEE 47th Annual International Symposium
  on Computer Architecture (ISCA)}}. IEEE, \bibinfo{pages}{446--459}.
\newblock


\bibitem[{ROCm}(2025)]%
        {ROCm:2025:HIP}
\bibfield{author}{\bibinfo{person}{{ROCm}}.} \bibinfo{year}{2025}\natexlab{}.
\newblock \bibinfo{title}{{HIP}: Heterogeneous Interface for Portability}.
\newblock \bibinfo{howpublished}{\url{https://github.com/ROCm/HIP}}.
\newblock
\newblock
\shownote{[Online; accessed 17-January-2025]}.


\bibitem[{ROCm Documentation Team}(2025)]%
        {ROCm:2025:PMD}
\bibfield{author}{\bibinfo{person}{{ROCm Documentation Team}}.}
  \bibinfo{year}{2025}\natexlab{}.
\newblock \bibinfo{title}{Performance Model}.
\newblock
  \bibinfo{howpublished}{\url{https://rocm.docs.amd.com/projects/rocprofiler-compute/en/latest/conceptual/performance-model.html}}.
\newblock
\newblock
\shownote{[Online; accessed 17-January-2025]}.


\bibitem[Team et~al\mbox{.}(2024)]%
        {Team:2024:GUU}
\bibfield{author}{\bibinfo{person}{Gemini Team}, \bibinfo{person}{Petko
  Georgiev}, \bibinfo{person}{Ving~Ian Lei}, \bibinfo{person}{Ryan Burnell},
  \bibinfo{person}{Libin Bai}, \bibinfo{person}{Anmol Gulati},
  \bibinfo{person}{Garrett Tanzer}, \bibinfo{person}{Damien Vincent},
  \bibinfo{person}{Zhufeng Pan}, \bibinfo{person}{Shibo Wang}, {et~al\mbox{.}}}
  \bibinfo{year}{2024}\natexlab{}.
\newblock \showarticletitle{{Gemini} 1.5: Unlocking Multimodal Understanding
  Across Millions of Tokens of Context}.
\newblock \bibinfo{journal}{\emph{arXiv preprint arXiv:2403.05530}}
  (\bibinfo{year}{2024}).
\newblock
\href{https://doi.org/10.48550/arXiv.2403.05530}{doi:\nolinkurl{10.48550/arXiv.2403.05530}}


\bibitem[Wang et~al\mbox{.}(2024)]%
        {wang2024preliminary}
\bibfield{author}{\bibinfo{person}{Zixian Wang}, \bibinfo{person}{Khai Vu},
  \bibinfo{person}{Miro Hodak}, \bibinfo{person}{Aarush Mehrotra},
  \bibinfo{person}{Francisco Gutierrez}, \bibinfo{person}{Kyle Smith},
  \bibinfo{person}{Gloria Seo}, \bibinfo{person}{Austin Garcia},
  \bibinfo{person}{Bryan Chin}, \bibinfo{person}{Marty Kandes}, {and}
  \bibinfo{person}{Mary~P Thomas}.} \bibinfo{year}{2024}\natexlab{}.
\newblock \showarticletitle{Preliminary Results of the MLPerf BERT Inference
  Benchmark on AMD Instinct GPUs}. In \bibinfo{booktitle}{\emph{Practice and
  Experience in Advanced Research Computing 2024: Human Powered Computing}}.
  \bibinfo{publisher}{Association for Computing Machinery}.
\newblock
\showISBNx{9798400704192}


\bibitem[Yang et~al\mbox{.}(2024)]%
        {Yang:2024:QTT}
\bibfield{author}{\bibinfo{person}{An Yang}, \bibinfo{person}{Baosong Yang},
  \bibinfo{person}{Beichen Zhang}, \bibinfo{person}{Binyuan Hui},
  \bibinfo{person}{Bo Zheng}, \bibinfo{person}{Bowen Yu},
  \bibinfo{person}{Chengyuan Li}, \bibinfo{person}{Dayiheng Liu},
  \bibinfo{person}{Fei Huang}, \bibinfo{person}{Haoran Wei}, {et~al\mbox{.}}}
  \bibinfo{year}{2024}\natexlab{}.
\newblock \showarticletitle{{Qwen}2.5 Technical Report}.
\newblock \bibinfo{journal}{\emph{arXiv preprint arXiv:2412.15115}}
  (\bibinfo{year}{2024}).
\newblock
\href{https://doi.org/10.48550/arXiv.2412.15115}{doi:\nolinkurl{10.48550/arXiv.2412.15115}}


\bibitem[Yang et~al\mbox{.}(2018)]%
        {Yang:2018:AER}
\bibfield{author}{\bibinfo{person}{Charlene Yang}, \bibinfo{person}{Rahulkumar
  Gayatri}, \bibinfo{person}{Thorsten Kurth}, \bibinfo{person}{Protonu Basu},
  \bibinfo{person}{Zahra Ronaghi}, \bibinfo{person}{Adedoyin Adetokunbo},
  \bibinfo{person}{Brian Friesen}, \bibinfo{person}{Brandon Cook},
  \bibinfo{person}{Douglas Doerfler}, \bibinfo{person}{Leonid Oliker},
  \bibinfo{person}{Jack Deslippe}, {and} \bibinfo{person}{Samuel Williams}.}
  \bibinfo{year}{2018}\natexlab{}.
\newblock \showarticletitle{An Empirical Roofline Methodology for
  Quantitatively Assessing Performance Portability}. In
  \bibinfo{booktitle}{\emph{2018 IEEE/ACM International Workshop on
  Performance, Portability and Productivity in HPC (P3HPC)}}.
  \bibinfo{pages}{14--23}.
\newblock
\href{https://doi.org/10.1109/P3HPC.2018.00005}{doi:\nolinkurl{10.1109/P3HPC.2018.00005}}


\bibitem[Zhu et~al\mbox{.}(2024)]%
        {Zhu:2024:DBC}
\bibfield{author}{\bibinfo{person}{Qihao Zhu}, \bibinfo{person}{Daya Guo},
  \bibinfo{person}{Zhihong Shao}, \bibinfo{person}{Dejian Yang},
  \bibinfo{person}{Peiyi Wang}, \bibinfo{person}{Runxin Xu},
  \bibinfo{person}{Y. Wu}, \bibinfo{person}{Yukun Li}, \bibinfo{person}{Huazuo
  Gao}, \bibinfo{person}{Shirong Ma}, {et~al\mbox{.}}}
  \bibinfo{year}{2024}\natexlab{}.
\newblock \showarticletitle{{DeepSeek-Coder-V2}: Breaking the Barrier of
  Closed-Source Models in Code Intelligence}.
\newblock \bibinfo{journal}{\emph{arXiv preprint arXiv:2406.11931}}
  (\bibinfo{year}{2024}).
\newblock
\href{https://doi.org/10.48550/arXiv.2406.11931}{doi:\nolinkurl{10.48550/arXiv.2406.11931}}


\end{thebibliography}
